\documentclass[letterpaper]{article} 
\usepackage{aaai2026}  
\usepackage{times}  
\usepackage{helvet}  
\usepackage{courier}  
\usepackage[hyphens]{url}  
\usepackage{graphicx} 
\usepackage{natbib}  
\usepackage{caption} 
\usepackage{algorithm}

\usepackage{multirow}
\usepackage{makecell}
\usepackage{booktabs}
\usepackage{amsmath}
\usepackage[table]{xcolor}

\usepackage[utf8]{inputenc}   
\usepackage[T1]{fontenc}      
\usepackage{CJKutf8}          
\usepackage{longtable}
\usepackage[most]{tcolorbox}
\usepackage{algpseudocode}

\usepackage{newfloat}
\usepackage{listings}
\DeclareCaptionStyle{ruled}{labelfont=normalfont,labelsep=colon,strut=off} 
\floatstyle{ruled}
\newfloat{listing}{tb}{lst}{}
\floatname{listing}{Listing}
\title{HSKBenchmark: Modeling and Benchmarking Chinese Second Language Acquisition in Large Language Models through Curriculum Tuning}
\author{
        Qihao Yang$^1$\equalcontrib,
    Xuelin Wang$^2$\equalcontrib,
    Jiale Chen$^1$,
    Xuelian Dong$^1$,
    Yuxin Hao$^{3\dagger}$,
    Tianyong Hao$^{1\dagger}$
}
\affiliations{

        \textsuperscript{\rm 1}School of Computer Science, South China Normal University, Guangzhou, China\\
    \textsuperscript{\rm 2}College of Chinese Language and Culture, Jinan University, Guangzhou, China\\
    \textsuperscript{\rm 3}School of Chinese Studies and Exchange, Shanghai International Studies University, Shanghai, China\\
    charlesyeung@m.scnu.edu.cn
}

\usepackage{bibentry}

\begin{document}

\maketitle
\renewcommand{\thefootnote}{$\dagger$} 
\footnotetext[2]{Corresponding authors.}
\renewcommand{\thefootnote}{} 
\renewcommand{\thefootnote}{\arabic{footnote}}

\begin{abstract}
Language acquisition is vital to revealing the nature of human language intelligence and has recently emerged as a promising perspective for improving the interpretability of large language models (LLMs). However, it is ethically and practically infeasible to conduct experiments that require controlling human learners' language inputs. This poses challenges for the verifiability and scalability of language acquisition modeling, particularly in Chinese second language acquisition (SLA). While LLMs provide a controllable and reproducible alternative, a systematic benchmark to support phase-wise modeling and assessment is still lacking. 
To address these issues, we propose HSKBenchmark, the first benchmark for staged modeling and writing assessment of LLMs in Chinese SLA. The benchmark covers HSK levels 3 to 6, comprising authentic textbooks with 6.76M tokens, 16K synthetic instruction data, 30 test topics and a linguistically-grounded evaluation system. To simulate human acquisition trajectories, a curriculum-tuning framework is introduced, which trains LLMs in a progression from beginner to advanced proficiency levels. Since language production in writing is a key perspective for observing SLA development, an evaluation system is established to probe LLMs in writing, including the coverage of level-based grammar items, writing errors, lexical complexity, syntactic complexity, and holistic scoring. We also develop an HSKAgent fine-tuned on 10K compositions from Chinese second language learners to automate this evaluation system.
Extensive experimental results demonstrate that HSKBenchmark not only models Chinese SLA effectively, but also serves as a reliable benchmark for dynamic writing assessment in LLMs. Our fine-tuned LLMs have writing performance on par with advanced human learners and exhibit human-like acquisition characteristics. The HSKBenchmark, HSKAgent, and checkpoints serve as foundational tools and resources, with the potential to pave the way for future research on language acquisition modeling and LLMs interpretability. Code and data are publicly available at: https://github.com/CharlesYang030/HSKB.
\end{abstract}

\begin{figure}[t!]
\centering
\includegraphics[width=1.0\columnwidth]{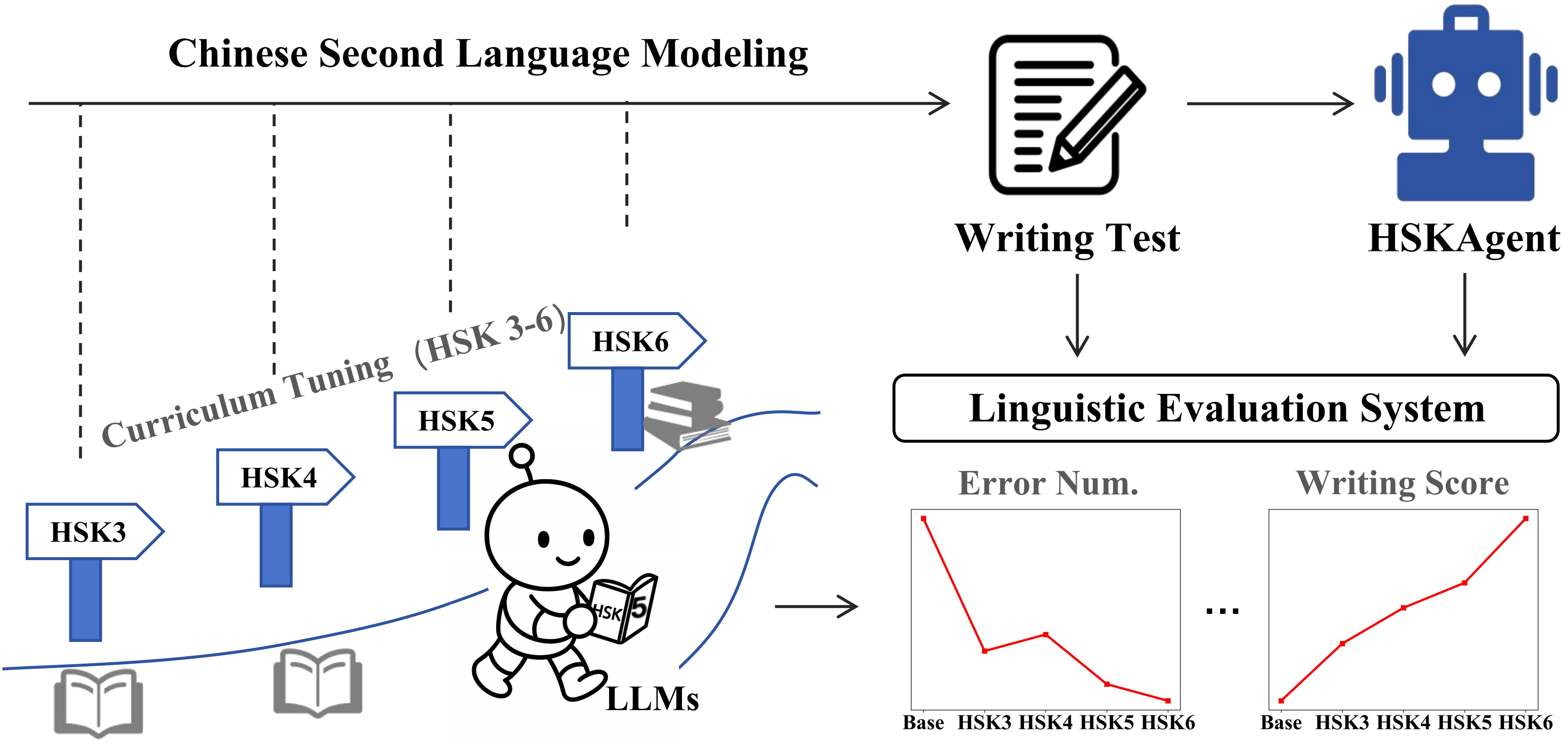} 
\caption{Illustration of Chinese SLA modeling and dynamic writing assessment in LLMs.}
\label{my_fig1}
\end{figure}


\section{Introduction}
Since the mid-20th century, research on language acquisition has advanced rapidly, laying theoretical foundations for understanding human language intelligence \cite{paper1,paper2,paper3}. However, due to ethical and practical limitations, many experiments involving controlled language inputs and the simulation of learning trajectories are difficult to conduct with human learners \cite{paper4}. As a result, the field has long faced challenges in terms of verifiability and computational modeling. Against this backdrop, large language models (LLMs) emerge as a valuable resource because of their controllability and reproducibility. The language acquisition of LLMs is receiving increasing attention. Researchers suggest that modeling the developmental patterns in LLMs not only enhances interpretability but also provides new theoretical and empirical insights into human language learning mechanisms \cite{paper6}.

Language acquisition is mainly categorized into first language (L1) acquisition and second language (L2) acquisition (SLA). Existing studies have explored L1 acquisition modeling of language models by adjusting the neural network architecture, optimizing hyperparameter settings, introducing linguistic features, or applying causal intervention \cite{paper4}. They achieve success in simulating children's vocabulary and grammar acquisition. Researchers attempt to transfer such success to SLA modeling. For example, a recent work trains XLM \cite{paper12} from scratch using a L1-L2 parallel corpus and observes that the model has similarities to humans in the transfer pattern from L1 to L2 \cite{paper11}. However, the SLA modeling of LLMs remains unresolved due to the lack of level-based training data and evaluation systems. Existing methods \cite{paper13} simply limit the size of training data rather than considering the difficulty of acquiring L2, resulting in unclear boundaries in SLA stages. Although different multilingual benchmarks are widely used to probe LLMs on various multilingual tasks, they mainly evaluate LLMs' existing capabilities \cite{paper14, paper26} rather than dynamic assessment for SLA modeling. Importantly, there are approximately 375 million English L2 learners
and 20 million Chinese L2 learners
in the world. The huge group stimulates an urgent need for empirical research on SLA modeling.

This paper studies an important yet overlooked issue: SLA modeling and dynamic writing assessment in LLMs, as shown in Figure \ref{my_fig1}. The \textbf{applicability of LLMs} is first considered: modeling SLA in LLMs requires selecting a non-English target language as L2, since most language models are trained primarily on large-scale English data. The \textbf{data accessibility} is also considered: there are extensive learning materials in Chinese, as \textit{Hanyu Shuiping Kaoshi} (HSK) \cite{paper40} is a representative Chinese L2 proficiency test. The \textbf{assessment method} is further considered: language production in writing is a key perspective for observing L2 development \cite{paper41}, which has advantages of reflecting the mastery of LLMs in the use of language structures. Based on these three considerations, in order to provide a reusable evaluation framework for SLA modeling, a feasible solution is to build a benchmark from the perspective of Chinese as L2 to assess the language output in writing of LLMs. Importantly, Chinese is an isolating language typologically distinct from English \cite{paper42}. Studying Chinese SLA modeling can be a representative view to examine whether LLMs can generalize across typologically diverse languages and capture structural patterns beyond Indo-European norms.

However, to achieve this goal, we encounter three major challenges. The first challenge is to build a benchmark with level-based training data. This requires using training data with clear level boundaries to distinguish acquisition stages developmentally, rather than merely controlling the scale of training data as in existing studies \cite{paper13,paper15}. The second challenge is to simulate human-like staged acquisition in LLMs and track its progression. This requires a curriculum-based design that incrementally exposes LLMs to staged Chinese inputs. The third challenge is to create an efficient evaluation system. This requires integrating linguistically-grounded indicators for LLMs writing and automating the system.

To address these challenges, we propose \textbf{HSKBenchmark}, the first benchmark for staged modeling and writing assessment of LLMs in Chinese SLA. To construct level-based training data, we collect 79 widely-used textbooks in international Chinese education, covering HSK levels 3 to 6. These textbooks with 6.76M tokens are used for staged pretraining. Following the \textit{Chinese Proficiency Grading Standards for International Chinese Language Education}, we identify 591 grammar items annotated with HSK levels. Three state-of-the-art LLMs (GPT, DeepSeek, Gemini) with robust Chinese capabilities are prompted to generate instruction data for writing exercises based on these grammar items. The 16k generated data is used for staged fine-tuning, with an agreement score of 0.91 and a validity rate of 95\%. In addition, thirty writing topics from real HSK exams are set as testing tasks. To simulate human-like staged acquisition, we introduce a curriculum-tuning framework, enabling LLMs to undergo staged pretraining followed by instruction tuning at each stage from HSK levels 3 to 6. For assessment, we build an evaluation system grounded in five linguistic dimensions: the coverage of level-based grammar items, writing errors, lexical complexity, syntactic complexity, and holistic scoring. We further develop an HSKAgent, an automated evaluator fine-tuned on the grammar dataset and 10K compositions from human Chinese L2 learners.

Our main contributions are summarized as follows:
\begin{itemize}
    \item The HSKBenchmark is proposed, which is the first benchmark for staged modeling and writing assessment of LLMs in Chinese SLA. It has the potential to serve as foundational tools and resources for future research on language acquisition modeling.
    \item A curriculum-tuning framework is introduced to simulate human language acquisition trajectories, and an HSKAgent is also developed to automate our linguistically-grounded evaluation system.
    \item Extensive experiments demonstrate the effectiveness of HSKBenchmark. Our fine-tuned LLMs achieve high writing performance on par with advanced human learners, contributing to the verification of SLA theories.
\end{itemize}

\begin{figure*}[t]
\centering
\includegraphics[width=1\textwidth]{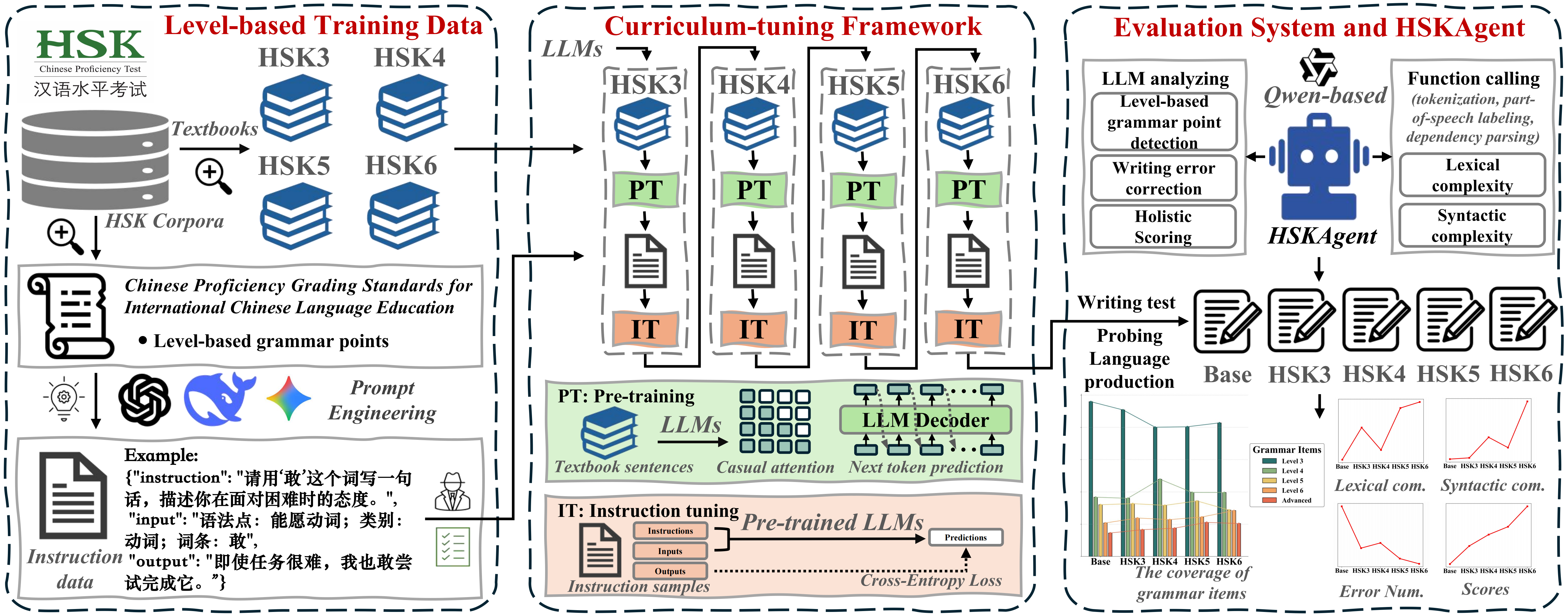} 
\caption{Illustration of our HSKBenchmark. It contains the level-based training data, the curriculum-tuning framework, the linguistically-grounded evaluation system and the HSKAgent.}
\label{my_fig2}
\end{figure*}

\section{Related Work}
\subsection{Language acquisition modeling with neural language models}
There has been much debate about the mechanism of language acquisition for a long time \cite{paper4}. To investigate the nature of language acquisition, neural language models were employed for language acquisition modeling in the 1980s \cite{paper50, paper16}. Although these early models had limited linguistic capabilities, their integration with cognitive science provided experimental insights into language mechanisms. In the past decade, with the advancement of natural language processing technology, language acquisition modeling has received renewed attention \cite{paper4}. While Dupre \cite{paper17} points out that language models lack real language learning capabilities, an increasing number of researchers believe they can be utilized as effective tools to verify language acquisition theories \cite{paper4, paper25}.

Existing work focuses mainly on modeling L1 acquisition \cite{paper4} to investigate the difference of inductive bias between human and machine \cite{paper19, paper20}. A recent work uses inductive bias distillation to transfer the Bayesian priors into the neural network \cite{paper21}. The research shows that such models not only learn languages from limited data, but also acquire complicated syntactic structures from large-scale corpora. Besides, many studies manipulate the internal structure of the models through controlling neural architectures \cite{paper7} and hyperparameters \cite{paper8}, or explore the structural bias of the models using linguistic features \cite{paper9} or causal interventions \cite{paper10}. A shared task named BabyLM \cite{paper22,paper23} was proposed recently to promote the development of evaluation frameworks for modeling child language acquisition.

In contrast, research on SLA modeling is still at an early stage and focuses primarily on L1–L2 transfer \cite{paper4,paper24}. A recent study explores the effects of L1-L2 transfer in the XLM model across different L1 (French, German, Russian, Japanese) and English as L2, finding that L1 pre-training significantly enhanced L2 syntactic generalization \cite{paper11}. The results indicate that transfer effects are influenced by typological distances and training configuration. However, such studies roughly distinguish the stages of language acquisition by controlling corpus size, lacking systematic modeling of the developmental trajectory of L2 production, especially in the context of Chinese as a second language \cite{paper13,paper15}. Therefore, this paper aims to adopt a curriculum-based approach and investigate the development of LLMs in linguistic competence in writing during the process of Chinese SLA modeling.

\subsection{Resources and evaluation in Chinese SLA}
The \textit{Hanyu Shuiping Kaoshi} (HSK) is currently the most widely-used standardized test to assess the Chinese proficiency of non-native Chinese learners \cite{paper40}. It consists of six levels (1 to 6) like Common European Framework of Reference for language (CEFR) \cite{paper27}, and provides a comprehensive evaluation of language skills including listening, speaking, reading, and writing. Many teaching resources are organized according to HSK levels, such as \textit{Developing Chinese} and \textit{Chinese Course}. In addition, open-access learner corpora like the \textit{HSK Dynamic Composition Corpus} contain manually annotated error corrections and proficiency scores. These materials offer a diverse and level-based source of training data for our work.

Linguistic complexity indices are widely used to evaluate the writing performance of Chinese L2 learners \cite{paper28,paper29,paper43}. The CTAP for Chinese \cite{paper30} achieves the automated extraction of 196 linguistic complexity indices across character, word, sentence, and paragraphs for Chinese learner writing. However, it does not calculate writing scores, which are a key indicator for measuring SLA development. While L2C-Rater \cite{paper31} predicts essay scores through regression models that integrate linguistic features, pre-extracted writing errors, and textual features, it lacks the ability to automatically detect errors for new compositions. Moreover, scoring essays through human teachers incurs high costs and low efficiency. Therefore, this paper aims to incorporate linguistic indicators that are specifically relevant to Chinese SLA development into the evaluation system, and to leverage LLMs with robust Chinese capabilities to develop an efficient agent for automated scoring.

\section{The HSKBenchmark}
To propose the HSKBenchmark, we make efforts from the construction of the level-based training data, the design of a curriculum-tuning framework, the development of a linguistically-grounded evaluation system and an HSKAgent, as shown in Figure \ref{my_fig2}.

\subsection{The construction of the level-based training data}
Krashen, one of the representative researchers in SLA research, argues that language acquisition occurs when learners are exposed incrementally to comprehensible input that contains linguistic features slightly beyond their current level (i+1) \cite{paper32}. In real L2 teaching scenarios, learners are also taught from beginner to advanced levels of teaching materials. However, existing studies do not pay attention to this issue because they usually distinguish the different stages of language acquisition based on the size of training data \cite{paper33, paper13}. For example, five learning stages can be divided in training data with 1 million tokens, where each batch of 200K tokens is regarded as one stage. In addition, the training data includes learning materials of different difficulties, without clearly distinguishing between beginner and advanced levels. To bridge this gap, we refer to the HSK level standard\footnote{Chinese learners in HSK level 3 can use Chinese to complete basic communication tasks in life, study, work, etc. Those in HSK level 6 can easily understand the Chinese information heard or read, and express their opinions fluently in Chinese in oral or written form. Detailed level-by-level descriptions can be accessed at: \url{https://www.chinesetest.cn/userfiles/file/dagang/HSK-koushi.pdf}.} that divides Chinese L2 proficiency into 6 levels, of which HSK levels 3 to 6 have writing tasks. Simultaneously, we conduct a survey of available resources for Chinese SLA. Two major issues are identified: (1) fewer learning materials are available at lower levels, particularly for HSK levels 1 and 2; (2) a substantial amount of manual effort is required to align multiple-choice questions in official HSK test collections with their corresponding answers, making it difficult to incorporate such items into the evaluation system like benchmarks in other domains.

To construct the level-based training data, we first collect 79 widely-used textbooks based on HSK levels 3 to 6, such as \textit{HSK Standard Course} and \textit{Boya Chinese Course}. These textbooks are a mixture of texts and images. We delete the images since multimodal inputs are not the objective of this study. In order to ensure the semantic compactness of the texts, we also delete all the Pinyin and English symbols used to assist learning in the textbooks through scripts. Finally, the total number of tokens in the cleaned textbooks is 6.76M, with 162,074 sentences and an average of 41.74 tokens per sentence, as shown in Table \ref{my_table1}.

\begin{table}[t]
\centering
\begin{tabular}{c|c|c|c}
    \toprule
    \textbf{\makecell[c]{Textbook\\levels}} & \textbf{Tokens} & \textbf{Sentences} & \textbf{\makecell[c]{Average number \\ of tokens \\per sentence}}\\
    \midrule
    HSK 3	&895,037&	22,743&	39.35\\
    HSK 4&	1,473,516&	34,637&	42.66\\
    HSK 5	&1,717,178	&41,044&	41.84\\
    HSK 6	&2,678,621	&63,650	&42.08\\
    Total	&6,764,352	&162,074	&41.74\\
    \bottomrule
\end{tabular}
\caption{Statistics of the level-based textbooks.}
\label{my_table1}
\end{table}

\begin{table}[t]
\centering
\setlength{\tabcolsep}{3pt}
\begin{tabular}{l|c|c|c|c|c|c}
    \toprule
    \textbf{Items} & \textbf{HSK3} & \textbf{HSK4} & \textbf{HSK5} & \textbf{HSK6} & \textbf{Advan.} & \textbf{Total}\\
    \midrule
    Word	&110	&48	&47	&50	&62&317 \\
    Phrase	&9	&6	&8&	11	&21&55\\
    FF&	5	&6	&6	&3&	5 & 25\\
    SC	&14	&4	&11	&3	&7&39\\
    ST	&27&	27	&26	&16	&47&143\\
    EU	&3	&4	&3	&1	&1&12\\
    ALL &168	&95	&101	&84	&143&591\\
    \midrule
    \midrule
    Num. & 4,600 & 2,607 & 2,896 & 2,334 & 4,025 &16,462\\
    \bottomrule
\end{tabular}
\caption{Statistics of the grammar items and the instruction data. Advan. refers to the advanced HSK level.}
\label{my_table2}
\end{table}

Besides textbooks, human teachers often ask learners to complete writing exercises to improve their language production ability. Therefore, we create a set of instruction data covering various writing exercises. Specifically, we first integrate HSK levels 3 to 6 and advanced grammar items from \textit{Chinese Proficiency Grading Standards for International Chinese Language Education}. Six types of grammar items are selected because they appear at these levels, including word, phrase, fixed format (FF), sentence component (SC), sentence type (ST), and emphatic usage (EU). Data that include multiple words or usages in the same grammar item are manually split. Secondly, we leverage GPT-4.1-mini \cite{paper34}, DeepSeek-Chat-V3 \cite{paper35}, and Gemini-2.5-Flash \cite{paper36} with robust Chinese capabilities to generate level-based instruction data according to these grammar items using in-context learning with two shots. The LLMs are prompted to generate 10 instruction instances for each grammar item. Each piece of generated data contains an instruction, an input, and an output (as shown in the example in Figure \ref{my_fig2}), where the instruction is the requirement of a writing exercise, the input is the specified grammar item, and the output is the expected language production. Then, three graduate annotators are recruited and trained on HSK standards. A randomly sampled set from the generated data is manually verified by the annotators using Fleiss's Kappa, yielding an agreement score of 0.91 and a validity rate of 95\%. Finally, we conduct proofreading and data filtering and then obtain 16,462 synthetic instruction data based on these 591 level-based grammar items. The statistics of the grammar items and the synthetic level-based instruction data are reported in Table \ref{my_table2}.

\subsection{The curriculum-tuning framework}
After distinguishing the stages in Chinese SLA using the level-based textbooks and instruction data, LLMs are also required to adapt to such staged modeling and assessment rather than being trained on all data at once. To this end, we introduce a curriculum-tuning framework, enabling LLMs to simulate Chinese L2 learners from self-learning on textbooks to writing exercises at each stage for gaining progressive capabilities in writing.

First, \textbf{pretraining on level-based textbooks for simulating input-based learning}: we define an HSK level \( l \in \{3, 4, 5, 6\} \), and the corresponding level-specific textbooks are denoted as \( \mathcal{T}^{(l)} = \{x_1, x_2, \ldots, x_m\} \), where each \( x_i \) is a Chinese sentence. A LLM adopts a causal language modeling architecture and is trained using next-token prediction to compute the loss for each sentence. The pretraining loss at level \( l \) is defined as:
\begin{equation}
\mathcal{L}_{\mathrm{PT}}^{(l)}=-\sum_{i=1}^m \sum_{t=1}^{\left|x_i\right|} \log P_{\theta^{(0)}}\left(x_{i, t} \mid x_{i<t}\right)
\end{equation}
where \( \theta^{(0)} \) denotes the LLM's initial parameters. For each sentence, \( x_{i,t} \) refers to its \( t \)-th token, and \( x_{i<t} \) denotes the preceding context before that token. After this stage of training, the resulting model is denoted as \( LLM-\theta_{\mathrm{PT}}^{(l)} \).

Second, \textbf{instruction tuning on writing exercises for simulating output-based learning}: we use the instruction data \( \mathcal{D}^{(l)} = \{(p_1, y_1), (p_2, y_2), \ldots, (p_n, y_n)\} \) corresponding to HSK level \( l \), where each \( p_i \) is a writing prompt and \( y_i \) is the target completion. In this paper, the writing prompt is the combination of the instruction (the requirement of writing exercises) and the input (the specific grammar item), and the completion is the output (the expected language production). The LLM is then fine-tuned on this instruction-following task using the same language modeling loss:
\begin{equation}
\mathcal{L}_{\mathrm{IT}}^{(l)}=-\sum_{i=1}^n \sum_{t=1}^{\left|y_i\right|} \log P_{\theta_{\mathrm{PT}}^{(l)}}\left(y_{i, t} \mid p_i, y_{i<t}\right)
\end{equation}
The resulting model after this stage of instruction tuning is denoted as \( LLM-\theta_{\mathrm{IT}}^{(l)} \).

Finally, \textbf{curriculum tuning across levels}: LLMs experience curriculum tuning in ascending order of levels, namely from HSK level 3 to 6. At each level \( l \), the LLM is first pretrained on the textbook data \( \mathcal{T}^{(l)} \) and then instruction-tuned on the corresponding instruction data \( \mathcal{D}^{(l)} \). The model parameters are updated at each level according to:
\begin{equation}
\theta_{\mathrm{PT}}^{(l)}=\operatorname{Pretraining}\left(\theta^{(l-1)}, \mathcal{T}^{(l)}\right)
\end{equation}
\begin{equation}
\theta_{\mathrm{IT}}^{(l)}=\operatorname{InstructionTuning}\left(\theta_{\mathrm{PT}}^{(l)}, \mathcal{D}^{(l)}\right)
\end{equation}
The final model \( LLM-\theta^{(6)} \) is obtained by sequential fine-tuning on all level-based textbooks and instruction data, thereby simulating a complete Chinese SLA trajectory.

\begin{table*}[t]
    \centering
    \setlength{\tabcolsep}{5pt}
    \begin{tabular}{l|c|c|c|c|c|c|c|c|c}
    \toprule
    \multirow{2}{*}{\textbf{Human/LLMs}} 
    & \multicolumn{5}{c|}{\small\textbf{\makecell[c]{The Coverage of Grammar Items}}} 
    & \textbf{\makecell[c]{Writing \\Errors}} 
    & \textbf{\makecell[c]{Lexical\\ Complexity}} 
    & \textbf{\makecell[c]{Syntactic \\Complexity}} 
    & \textbf{\makecell[c]{Holistic \\Scoring}} \\
    \cline{2-10}
    ~ & HSK3  & HSK4 & HSK5 & HSK6 & Advan. & Err & MATTR-50 & MDD & Score \\
    \midrule
Natives & 0.3408 & 0.2439 & 0.1745 & 0.1261 & 0.1146 & 1.4000 & 0.8061 & 2.9769 & 88.3333\\
Leaner-95* & 0.3563 & 0.2040 & 0.1656 & 0.1392 & 0.1350 & \textbf{2.8667} & \textbf{0.8165} & \textbf{2.8386} & \textbf{85.0000}\\
Leaner-90* & 0.3481 & 0.1854 & 0.1997 & 0.1425 & 0.1243 & \textbf{3.3667} & \textbf{0.8059} & \textbf{2.9705} & \textbf{84.0000}\\
Leaner-80* & 0.3855 & 0.1914 & 0.1835 & 0.1327 & 0.1069 & \textbf{3.5000} & \textbf{0.7925} & \textbf{2.6473} & \textbf{74.8333}\\
Leaner-70* & 0.3802 & 0.2094 & 0.1978 & 0.1211 & 0.0915 & \textbf{3.8333} & \textbf{0.7764} & \textbf{2.6205} & \textbf{70.1667}\\
Leaner-60* & 0.3947 & 0.2030 & 0.1967 & 0.1034 & 0.1021 & \textbf{4.8000} & \textbf{0.7806} & \textbf{2.5814} & \textbf{63.0000}\\
GPT-4.1-mini & 0.3979 & 0.2324 & 0.1622 & 0.1082 & 0.0993 & 0.0000 & 0.8287 & 2.6032 & 91.5000\\
DeepSeek-Chat & 0.4102 & 0.2118 & 0.1615 & 0.1166 & 0.0999 & 0.0000 & 0.8427 & 2.5411 & 92.3333\\
Gemini-2.5 & 0.4038 & 0.2265 & 0.1673 & 0.1103 & 0.0921 & 0.0000 & 0.8334 & 2.5894 & 90.5300\\
\midrule
Llama2 & 0.4844 & 0.1615 & 0.1667 & 0.1126 & 0.0748 & 0.9000 & 0.6860 & 2.4253 & 70.0000\\
Llama2$_{\mathrm{HSK3}}$ & \textbf{0.4925$\uparrow$} & 0.1738 & 0.1471 & 0.1143 & \textbf{0.0723$\downarrow$} & \textbf{0.6333$\downarrow$} & \textbf{0.7188$\uparrow$} & \textbf{2.5045$\uparrow$} & \textbf{75.8333$\uparrow$}\\
Llama2$_{\mathrm{HSK4}}$ & 0.4517 & \textbf{0.2048$\uparrow$} & 0.1768 & 0.0880 & \textbf{0.0787$\uparrow$} & \textbf{0.6667$\downarrow$} & \textbf{0.7364$\uparrow$} & \textbf{2.5503$\uparrow$} & \textbf{78.6667$\uparrow$}\\
Llama2$_{\mathrm{HSK5}}$ & 0.4203 & 0.2005 & \textbf{0.1852$\uparrow$} & 0.1111 & \textbf{0.0829$\uparrow$} & \textbf{0.5667$\downarrow$} & \textbf{0.7592$\uparrow$} & \textbf{2.5274$\downarrow$} & \textbf{80.6667$\uparrow$}\\
Llama2$_{\mathrm{HSK6}}$ & 0.4246 & 0.1818 & 0.1775 & \textbf{0.1279$\uparrow$} & \textbf{0.0883$\uparrow$} & \textbf{0.5333$\downarrow$} & \textbf{0.7641$\uparrow$} & \textbf{2.5558$\uparrow$} & \textbf{81.8333$\uparrow$}\\
\midrule
Ch-Alpaca & 0.4470 & 0.2000 & 0.1678 & 0.1191 & 0.0661 & 0.0667 & 0.7705 & 2.5251 & 77.5000\\
Ch-Alpaca$_{\mathrm{HSK3}}$ & \textbf{0.4270$\downarrow$} & 0.1917 & 0.1803 & 0.1105 & \textbf{0.0905$\uparrow$} & \textbf{0.5000$\downarrow$} & \textbf{0.7774$\uparrow$} & \textbf{2.5329$\uparrow$} & \textbf{75.8333$\downarrow$}\\
Ch-Alpaca$_{\mathrm{HSK4}}$ & 0.4049 & \textbf{0.2252$\uparrow$} & 0.1639 & 0.1069 & \textbf{0.0990$\uparrow$} & \textbf{0.0333$\downarrow$} & \textbf{0.7726$\downarrow$} & \textbf{2.5109$\downarrow$} & \textbf{82.0000$\uparrow$}\\
Ch-Alpaca$_{\mathrm{HSK5}}$ & 0.3980 & 0.1859 & \textbf{0.2146$\uparrow$} & 0.1250 & \textbf{0.0765$\downarrow$} & \textbf{0.1000$\downarrow$} & \textbf{0.7816$\uparrow$} & \textbf{2.5557$\uparrow$} & \textbf{87.6667$\uparrow$}\\
Ch-Alpaca$_{\mathrm{HSK6}}$ & 0.3844 & 0.2382 & 0.1632 & \textbf{0.1161$\downarrow$} & \textbf{0.0981$\uparrow$} & \textbf{0.0000$\downarrow$} & \textbf{0.7829$\uparrow$} & \textbf{2.5729$\uparrow$} & \textbf{85.6667$\uparrow$}\\
\midrule
Mistral & 0.4798 & 0.1836 & 0.1603 & 0.1037 & 0.0726 & 0.7333 & 0.5260 & 2.5302 & 76.8333\\
Mistral$_{\mathrm{HSK3}}$ & \textbf{0.4542$\downarrow$} & 0.1802 & 0.1637 & 0.1190 & \textbf{0.0829$\uparrow$} & \textbf{0.5667$\downarrow$} & \textbf{0.7566$\uparrow$} & \textbf{2.5334$\uparrow$} & \textbf{79.5000$\uparrow$}\\
Mistral$_{\mathrm{HSK4}}$ & 0.4006 & \textbf{0.2393$\uparrow$} & 0.1583 & 0.1141 & \textbf{0.0876$\uparrow$} & \textbf{0.4667$\downarrow$} & \textbf{0.7788$\uparrow$} & \textbf{2.5858$\uparrow$} & \textbf{81.1667$\uparrow$}\\
Mistral$_{\mathrm{HSK5}}$ & 0.4020 & 0.1983 & \textbf{0.1719$\uparrow$} & 0.1222 & \textbf{0.1056$\uparrow$} & \textbf{0.3667$\downarrow$} & \textbf{0.7901$\uparrow$} & \textbf{2.5595$\downarrow$} & \textbf{82.3333$\uparrow$}\\
Mistral$_{\mathrm{HSK6}}$ & 0.4141 & 0.1981 & 0.1437 & \textbf{0.1422$\uparrow$} & \textbf{0.1019$\downarrow$} & \textbf{0.3000$\downarrow$} & \textbf{0.7886$\downarrow$} & \textbf{2.6772$\uparrow$} & \textbf{85.3333$\uparrow$}\\

    \bottomrule
    \end{tabular}
    \caption{The Chinese SLA performance of human and LLMs on HSKBenchmark. Learners-\textit{X}* refers to those who got a original score of \textit{X} in the \textit{HSK Dynamic Composition Corpus v2.0}. Ch-Alpaca indicates the Chinese-Alpaca model. The upward and downward arrows indicate whether the model's current performance has improved or declined compared to its previous level.}
    \label{my_table3}
\end{table*}
\subsection{The linguistically-grounded evaluation system and an HSKAgent}
To fairly evaluate the writing performance of LLMs, we collect 30 writing topics from the \textit{HSK Dynamic Composition Corpus v2.0}\footnote{\url{https://yuyanziyuan.blcu.edu.cn/info/1043/1501.htm}} as test tasks. This corpus, released by Beijing Language and Culture University, is a collection of written compositions produced by non-native Chinese speakers from 85 countries (32.85\% from Korea) in HSK test from 1992 to 2005. It includes more than 10K compositions with 4 million Chinese characters. These selected 30 topics cover a range of genres (e.g., narrative and argumentative writing) and topics (e.g., daily life and study). After examination, there is no data overlap or contamination between these 30 topics and our training data.

To capture and reflect the development of Chinese SLA across levels, we design an evaluation system by following previous work, covering five linguistic dimensions. (1)\textit{\textbf{The Coverage of Grammar Items}} refers to the proportion of grammar items from each HSK level in compositions. This metric is used to evaluate LLMs' mastery of grammar items across different proficiency levels. (2)\textit{\textbf{Writing Errors}} (\textbf{Err}) \cite{paper53} refers to the sum of character-level errors, lexical errors, syntatic errors and discourse-level errors. This metric is used to evaluate the accuracy of LLMs' language output. (3)\textit{\textbf{Lexical Complexity}} (\textbf{MATTR-50}) \cite{paper55} refers to the ratio of word types to word tokens within text windows, where each batch of 50 tokens is set as one window. This metric is used to evaluate LLMs' lexical proficiency. (4)\textit{\textbf{Syntatic Complexity}} (\textbf{MDD})  \cite{paper52} refers to the average dependency distance of texts. This metric is used to evaluate LLMs' syntactic proficiency. A higher MDD indicates longer dependency relations, which may reflect more sophisticated sentence structures. (5)\textit{\textbf{Holistic Scoring}} (\textbf{Score}) \cite{paper54} refers to the overall score, which is typically determined based on the length, quality and the relevance of the text.

To automate the evaluation system, we develop an \textbf{HSKAgent} built upon Qwen3-8B \cite{paper48}. The Qwen3-8B model is selected due to its strong performance in Chinese among 7/8B-scale models based on the SuperCLUE leaderboard\footnote{\url{https://www.superclueai.com/}}. It also has advantages in reproducibility and inference efficiency. Specifically, we transform the level-based instruction data into a binary classification dataset. For the positive samples, the original prompt and completion are concatenated into a new positive prompt, with the corresponding answer ``Yes". For the negative samples, the prompt is paired with a negative completion randomly sampled from the data pool, resulting in a new negative prompt with the answer ``No". To reduce the likelihood that the negative completion still aligns with the target grammar item, we restrict sampling to completions outside the current grammar item category. Although this is a straightforward approach, a manual validation yields an inter-annotator agreement score of 0.93 and a validity rate of 96\%. Then, we reconstruct the original human-written versions from these 10K compositions with error annotations and scores. This dataset is used to train and test the HSKAgent. Eventually, our HSKAgent achieves an F1-score of 0.97 for binary classification of grammar items, 90\% accuracy for error detection, and an F1-score of 0.81 for holistic scoring. It also obtains good agreements with human raters (Quadratic Weighted Kappa (QWK) = 0.7969, Spearman = 0.8010, Pearson = 0.8023). For complexity-related indices, the HSKAgent leverages function calling for automatic computation.

\section{Experiments and Results}
\subsection{Implementation details}
\textbf{Baselines.} Since our objective is not to train LLMs to acquire Chinese from scratch, we select LLMs that already possess a certain degree of Chinese capabilities, to investigate their developments during the Chinese SLA modeling. Therefore, we refer to SuperCLUE and choose three models of relatively low rank as baselines, including LLaMA2-7B-Chat, Mistral-7B-Instruct-v0.3, and Chinese-Alpaca-2-7B. Three stronger LLMs, GPT-4.0-mini, DeepSeek-Chat-V3, and Gemini-2.5-Flash, are also selected as baselines. Moreover, we include Chinese native speakers and Chinese L2 learners as human baselines.\\
\textbf{Setting.} The experiments are implemented on PyTorch 2.6.0 and 3 RTX 3090 GPUs (24GB) using LLaMA-Factory \cite{paper38}. LoRA \cite{paper44} is utilized to fine-tune these LLMs and the HSKAgent in pretraining and instruction tuning, where the learning rate is 5e-5, the number of epoch is 3 and bf16 is used as the compute type.

\subsection{Main results}
The Chinese SLA performance of human and LLMs on HSKBenchmark is reported in Table \ref{my_table3}. Compared with Chinese SLA learners, the natives achieve the highest overall score (88.3333). Although the advanced learners (95* and 90*) also score more than 80, there is still a noticeable gap between them and the natives in terms of writing errors and syntactic complexity. Moreover, as learners improve their proficiency from 60* to 95*, their scores also gradually increase, which provides evidence that there is indeed a predictable developmental progress in Chinese SLA and our HSKAgent indeed presents such a trend reasonably. GPT, DeepSeek, and Gemini obtain average scores exceeding 90, but they are inferior to humans in syntactic complexity and mastery of advanced grammar items.

LLaMA2, Chinese-Alpaca, and Mistral all exhibit substantial improvements after Chinese SLA modeling. For example, the base LLaMA2 model achieves a score of only 70, roughly equivalent to that of Learners-70*. After modeling at HSK3, LLaMA2$_{\mathrm{HSK3}}$ improves by 5.83 points, and the final LLaMA2$_{\mathrm{HSK6}}$ achieves a score of 81.83 on par with Learners-90*. In addition, the coverages of HSK3 and HSK4 grammars of LLaMA2$_{\mathrm{HSK3}}$ are 49.25\% and 17.38\%, but LLaMA2$_{\mathrm{HSK4}}$ shows a 4.08\% decrease and a 3.10\% increase respectively in these two aspects. This indicates that the curriculum-tuning framework enables the model to better acquire more complex grammars. LLMs with HSK5 and HSK6 levels get a higher proportion of advanced grammar items that are not included in training data, compared with those LLMs with HSK3 and HSK4 levels. This suggests that more advanced models may develop emergent abilities to master higher-level grammars and generalize beyond the training data, much like the human capacity to infer and extend learned knowledge. 


Compared with human learners, LLMs are less prone to produce errors. A possible reason is that the language production mechanisms in writing of humans and LLMs are fundamentally different. Compared with LLMs, human writers might tend to take more risks in those usages they are not fully confident in. Limited by the top-k next-token prediction mechanism, LLMs tend to generate only those tokens in which they have the highest confidence. However, LLMs fall short of humans in lexical and syntactic complexity. LLMs optimize for predictive likelihood, tending to generate shorter, more typical sentences found in natural corpora. In contrast, L2 learners often deliberately use complex structures in writing tests to display linguistic competence, leading to higher syntactic complexity. 

In summary, Table \ref{my_table3} presents comparisons between native speakers and L2 learners, between humans and LLMs, as well as the developmental trajectories of baseline LLMs in Chinese SLA. These results are consistent with expectations and support the effectiveness of our HSKBenchmark as an effective suite for benchmarking Chinese SLA performance.

\begin{figure}[t!]
\centering
\includegraphics[width=1.0\columnwidth]{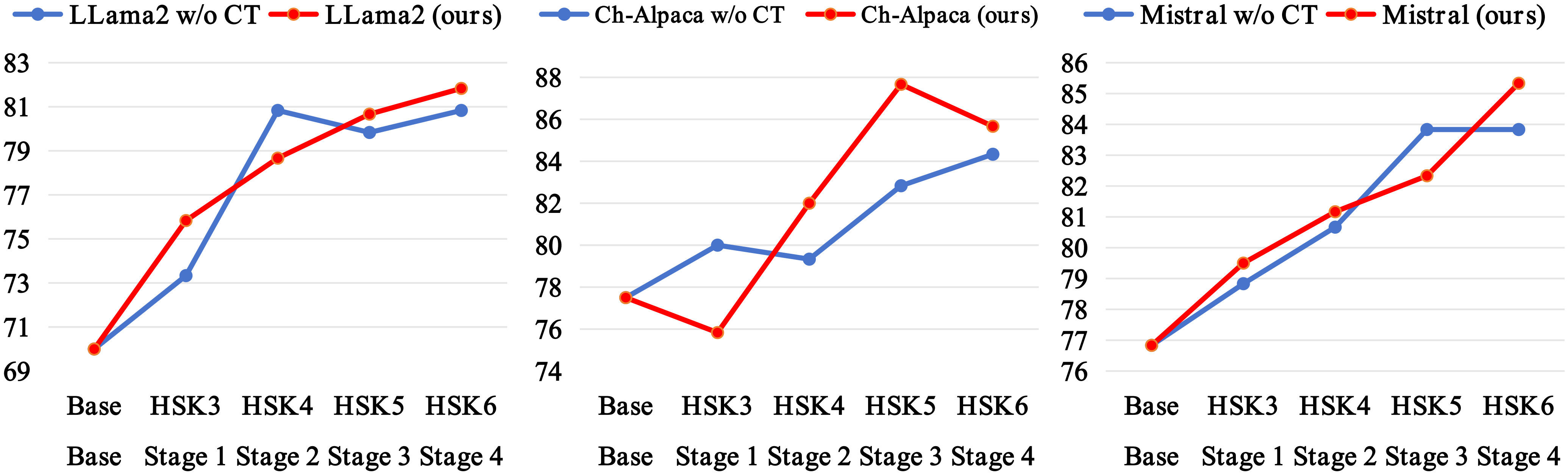} 
\caption{Comparison between our LLMs and those trained on the shuffled dataset in overall average scores. CT refers to the curriculum tuning.}
\label{my_fig3}
\end{figure}

\subsection{Ablation study}
An ablation study is conducted to reveal the effectiveness of our curriculum-tuning framework. Specifically, we shuffle and merge all level-based textbooks and instruction data into a single dataset. It is then divided into four stages (corresponding to HSK levels 3 to 6) purely based on data volume. The LLMs are finetuned on this dataset without level-based ordering. Figure \ref{my_fig3} illustrates the comparison between our LLMs trained on the curriculum-tuning framework and those trained on the shuffled pretraining method in overall average scores. The results show that the shuffled approach enables LLMs to achieve relatively higher average scores in the early stages, likely because the models are exposed to high-level training data prematurely. However, in the later stages (stage 3-4), the performance of our LLMs surpasses that of the shuffled approach. This suggests that even when trained on the same data, an appropriate learning sequence is essential for activating better Chinese SLA outcomes in LLMs. This finding not only validates the effectiveness of our curriculum-tuning framework, but also aligns with Krashen’s i+1 input hypothesis \cite{paper32}. This is because that our HSKBenchmark provides the training data with progressive difficulties like the i+1 input hypothesis which emphasizes the importance of progressively structured input in successful L2 acquisition.

\begin{figure}[t!]
\centering
\includegraphics[width=1\columnwidth]{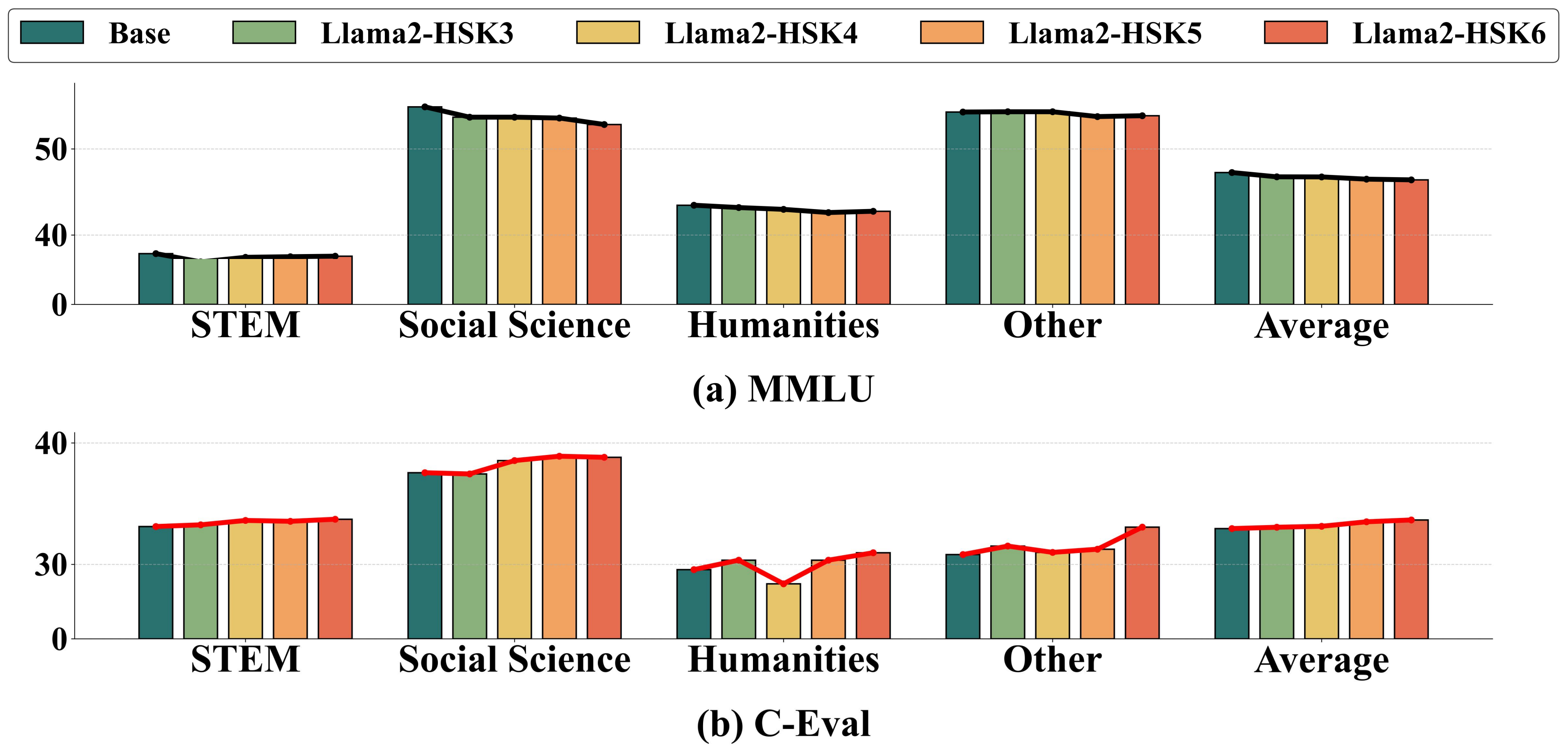} 
\caption{The performance of Llama2 on MMLU and C-Eval.}
\label{my_fig4}
\end{figure}

\subsection{Impact on L1 proficiency and general Chinese performance}
An additional experiment is conducted to investigate whether other language capabilities of LLMs change in their L1 proficiency and general Chinese performance during the process of Chinese SLA modeling. Llama2 is selected to be evaluated on two open-sourced benchmarks, MMLU \cite{paper14} and C-Eval \cite{paper39}. MMLU is a widely-used multitask English benchmark with QAs in STEM, social science, humanities and other subjects. C-Eval is a widely-used comprehensive Chinese exam benchmark with similar QAs. The results, as shown in Figure \ref{my_fig4}, show that Llama2 does not suffer degradation in L1 performance (no catastrophic forgetting) on MMLU and even exhibits slight L2 improvements on C-Eval. This pattern is similar to the behavior of human L2 learners, suggesting that the curriculum-tuned LLMs trained on HSKBenchmark present human-like characteristics.  This finding might support extending our method to other language frameworks (such as CEFR) to uncover more empirical insights about SLA modeling.

\section{Conclusion}
This paper proposes HSKBenchmark, the first benchmark for staged modeling and writing assessment of LLMs in Chinese SLA. A curriculum-tuning framework is introduced to simulate human language acquisition trajectories. A linguistically-grounded evaluation system is designed to assess the language production of LLMs in writing, and an HSKAgent is developed to automate the evaluation system. Experimental results demonstrate that HSKBenchmark effectively supports Chinese SLA modeling in LLMs. The curriculum-tuning framework facilitates more robust SLA development compared to traditional training approaches, and the evaluation system and HSKAgent successfully capture and reflect this developmental progress. The suite of models developed in this work will be released to serve as effective tools and resources for the community. In future work, we will scale the SLA modeling framework to a broader range of languages, incorporate multimodal inputs, and integrate additional linguistic dimensions to further explore the potential of LLMs in computational modeling and advancing SLA theories.

\section{Acknowledgement}
The work was supported by grants from National Natural Science Foundation of China (No. 62372189), the Research Grants Council of the Hong Kong Special Administrative Region, China (UGC/FDS16/E03/25), Fujian Provincial Social Science Foundation Project (No. FJ2025B104) and ``the Fundamental Research Funds for the Central Universities", and the China Scholarship Council (No. 202406780045).

\bibliography{main}

\newpage
\section{Appendix}
The appendix provides supplementary materials for the main paper, including the detailed descriptions about the textbooks, the grammar items, the prompt engineering for LLMs to generate instruction data, the pseudocode for the curriculum-tuning framework, the binary classification dataset, the error detection, and the HSKAgent platform.

\section{A. Textbook checklists}
To construct the level-based training data, we first collect 79 widely-used textbooks based on HSK levels 3 to 6, such as \textit{HSK Standard Course} and \textit{Boya Chinese Course}. These textbooks are a mixture of texts and images. We manually delete the images since the multimodal inputs are not the objective of this paper. In order to ensure the semantic compactness of the texts, we also delete all the Pinyin and English symbols used to assist learning in the textbooks through scripts. Finally, the total number of tokens in the cleaned textbooks is 6.76M, with 162,074 sentences and an average of 41.74 tokens per sentence.

In addition, we conduct a systematic analysis of these textbooks. The checklist of textbooks corresponding HSK levels 3 and 4 is presented in Table \ref{appendix_table1}. The checklist of textbooks corresponding HSK levels 5 and 6 is presented in Table \ref{appendix_table2}. Overall, the higher the proficiency level, the fewer textbooks are available. Note that some textbooks did not have official English titles at the time of publication. For these textbooks, we provide translated English titles for reference. These translations may not be entirely accurate. Therefore, it is recommended to use the original Chinese titles when searching for the source materials.

\section{B. Checklists of the level-based grammar items}
To create a set of instruction data covering various writing exercises, we first integrate HSK levels 3 to 6 and advanced grammar items from \textit{Chinese Proficiency Grading Standards for International Chinese Language Education}. According to the information retrieval system\footnote{\url{https://old.chinesetest.cn/standardsAction.do?means=standardInfo}} from Chinese Tests Service Website,  six types of grammar items are selected because they appear at these levels, including word, phrase, fixed format (FF), sentence component (SC), sentence type (ST), and emphatic usage (EU). The checklist of grammar items with HSK level 3 is shown in Table \ref{appendix_mytable3}. The checklist of grammar items with HSK level 4 is shown in Table \ref{appendix_mytable4}. The checklist of grammar items with HSK level 5 is shown in Table \ref{appendix_mytable5}. The checklist of grammar items with HSK level 6 is shown in Table \ref{appendix_mytable6}. The checklist of grammar items with HSK advanced level is shown in Table \ref{appendix_mytable7}.

\begin{algorithm}[t]
\caption{Pseudocode of curriculum tuning}
\label{pseudocode}
\begin{algorithmic}[1]
\State \textbf{Input:} Base model $\theta_0$, HSK levels $\mathcal{L} = \{3,4,5,6\}$
\For{level $l \in \mathcal{L}$}
    \State TextbookData[$l$] $\gets$ list of sentences for level $l$
    \State InstructionData[$l$] $\gets$ list of (prompt, completion) pairs for level $l$
\EndFor
\State $\theta \gets \theta_0$ \Comment{Initialize base model}

\For{level $l \in \mathcal{L}$}
    \ForAll{sentence $\in$ TextbookData[$l$]}
        \State $loss_{pt} \gets \text{PretrainingLoss}(\theta, \text{sentence})$
        \State $\theta \gets \text{UpdateModel}(\theta, loss_{pt})$
    \EndFor
    \ForAll{(prompt, completion) $\in$ InstructionData[$l$]}
        \State $loss_{it} \gets \text{InstructionLoss}(\theta, \text{prompt}, \text{completion})$
        \State $\theta \gets \text{UpdateModel}(\theta, loss_{it})$
    \EndFor
\EndFor

\State \Return $\theta$ \Comment{Final $LLM-\theta^{(6)}$ after curriculum tuning}
\end{algorithmic}
\end{algorithm}

\section{C. Prompting LLMs to generate instruction data}
We leverage GPT-4.1-mini \cite{paper34}, DeepSeek-Chat-V3 \cite{paper35}, and Gemini-2.5-Flash \cite{paper36} with robust Chinese capabilities to generate level-based instruction data according to these grammar items using in-context learning with two shots. The LLMs are prompted to generate 10 instruction instances for each grammar item. Each piece of generated data contains an instruction, an input, and an output, where the instruction is the requirement and background of the writing exercise, the input is the specified grammar item, and the output is the expected language production. The prompt, as shown in Figure \ref{appendix_myfig1}, is used to instruct the three LLMs to generate instruction data based on these grammar items. 

\section{D. Training LLMs through the curriculum-tuning framework}
Llama2-7B-Chat, Mistral-7B-Instruct-v0.3, and Chinese-Alpaca-2-7B are selected as baselines to be trained using the curriculum-tuning framework. The framework contains three process: pretraining on level-based textbooks for simulating input-based learning, instruction tuning on writing exercises for simulating output-based learning, curriculum tuning across levels. The pseudocode of this training process is abstracted in Algorithm \ref{pseudocode}.

\section{E. The binary classification dataset for grammar detection}
To enable the HSKAgent to detect grammar items from compositions, a binary classification dataset is constructed by transforming the level-based instruction data into positive data and negative data. To reduce the likelihood that the negative completion still aligns with the target grammar item, we restrict sampling to completions drawn from outside the current grammar item category. For the positive samples, the original prompt and completion are concatenated into a new positive prompt, with the corresponding answer ``Yes". A positive example is shown as follow:\\
\begin{CJK}{UTF8}{gbsn}       
-----------------------------------------------------------------------\\
\hspace*{1em} \{

\hspace*{1em}"\textbf{instruction}": "你是一个HSK语法点检测器，请你对以下句子进行判断是否符合当前语法点：No:133, 级别:3级, 语法项目:词类, 类别:动词, 细目:能愿动词, 语法内容:敢。",

\hspace*{1em}"\textbf{input}": "句子：即使任务很难，我也敢尝试完成它。",

\hspace*{1em}"\textbf{output}": "是",\\
\hspace*{1em} \}\\
-----------------------------------------------------------------------
\end{CJK}\\
For the negative samples, the prompt is paired with a negative completion randomly sampled from the data pool, resulting in a new negative prompt with the answer ``No". A negative example is shown as follow:\\
\begin{CJK}{UTF8}{gbsn}       
-----------------------------------------------------------------------\\
\hspace*{1em} \{

\hspace*{1em}"\textbf{instruction}": "你是一个HSK语法点检测器，请你对以下句子进行判断是否符合当前语法点：No:133, 级别:3级, 语法项目:词类, 类别:动词, 细目:能愿动词, 语法内容:敢。",

\hspace*{1em}"\textbf{input}": "句子：我最喜欢春节，它不仅仅是鞭炮声声的热闹，是家家户户张灯结彩的喜庆，更是它背后所承载的团圆文化是深厚的，辞旧迎新的寓意是美好的。我对春节的喜爱是如此强烈，因为它带来的，是家的温暖，是年的味道，是所有中国人共同的记忆。",

\hspace*{1em}"\textbf{output}": "否",\\
\hspace*{1em} \}\\
-----------------------------------------------------------------------
\end{CJK}

Following this pipeline, the binary classification dataset contains an equal number of positive and negative data, which is then used to finetune the HSKAgent.

\section{F. The error detection of HSKAgent}
Error detection is one of our HSKAgent's functions for evaluating the accuracy of LLMs' language production in writing during the process of Chinese SLA. After finetuning, the HSKAgent is able to detect errors for new compositions. An example is shown as follow:\\
\begin{CJK}{UTF8}{gbsn}       
-----------------------------------------------------------------------\\
\textbf{The original content of a composition:}

\hspace*{1em}我从小到大，对我影响最大的一个人是我的母亲。我的母亲是典形家庭妇女。但我认为在这个世界上最伟大的人。不因为她生了我，才说最伟大而是客观的角度上说是最伟大。我母亲是个典形乡下人。已经在大城市里住了三十年，还带着土味对待城市人。所以很多人叫她‘城市的乡下老儿’。她也很喜欢听别人这样称呼她。但我呢？倒也不喜欢听这个别名。因为一听这个别名就觉得他们看不起我的母亲。我知道母亲是多么健强的人。我从小开始她教我知道什么是自立和独立，什么是叫责任感。这都是她自己的语言和行动来教我。可以说是那时开始学了怎么做好的人。她让我深刻地懂‘责任感’。有一天把我打得 (end)\\
\textbf{The version for error detection and correction:}

我从小到大，对我影响最大的一个人是我的母亲。我的母亲是典型[B形]\{CQ的\}家庭妇女，[BC。]但我认为在这个世界上\{CQ她是\}最伟大的人。不只是\{CC因为\}她生了我，才说最伟大\{CQ的\}[BQ，]而是\{CQ从\}客观的角度上说是最伟大\{CQ的\}。我母亲是个典型[B形]\{CQ的\}乡下人，[BC。]已经在大城市里住了三十年，还带着土味对待城市人。所以很多人叫她“[BC‘]城市的乡下佬[B老]儿”[BC’]，[BC。]她也很喜欢听别人这样称呼她。但我呢？倒\{CD也\}不喜欢听这个别名。因为一听\{CQ到\}这个别名就觉得他们看不起我的母亲。我知道母亲是多么坚[B健]强的人。我从小开始[BQ，]她\{CQ就\}教我知道什么是自立和独立，什么是\{CD叫\}责任感。这些\{CC2这\}都是她\{CQ用\}自己的语言和行动来教我。可以说是那时\{CQ我\}开始学习\{CC学了\}怎么做好的人。她让我深刻地懂\{CQ得\}“[BC‘]责任感”[BC’]。有一天把我打得\{WWJ\}\\
-----------------------------------------------------------------------
\end{CJK}

Based on the collected compositions written by human L2 learners, all error types and an error handling guideline are published in our GitHub repository.

\section{G. The platform of HSKAgent}
\begin{figure}[H]
\centering
\fcolorbox{gray}{white}{\includegraphics[width=0.7\textwidth]{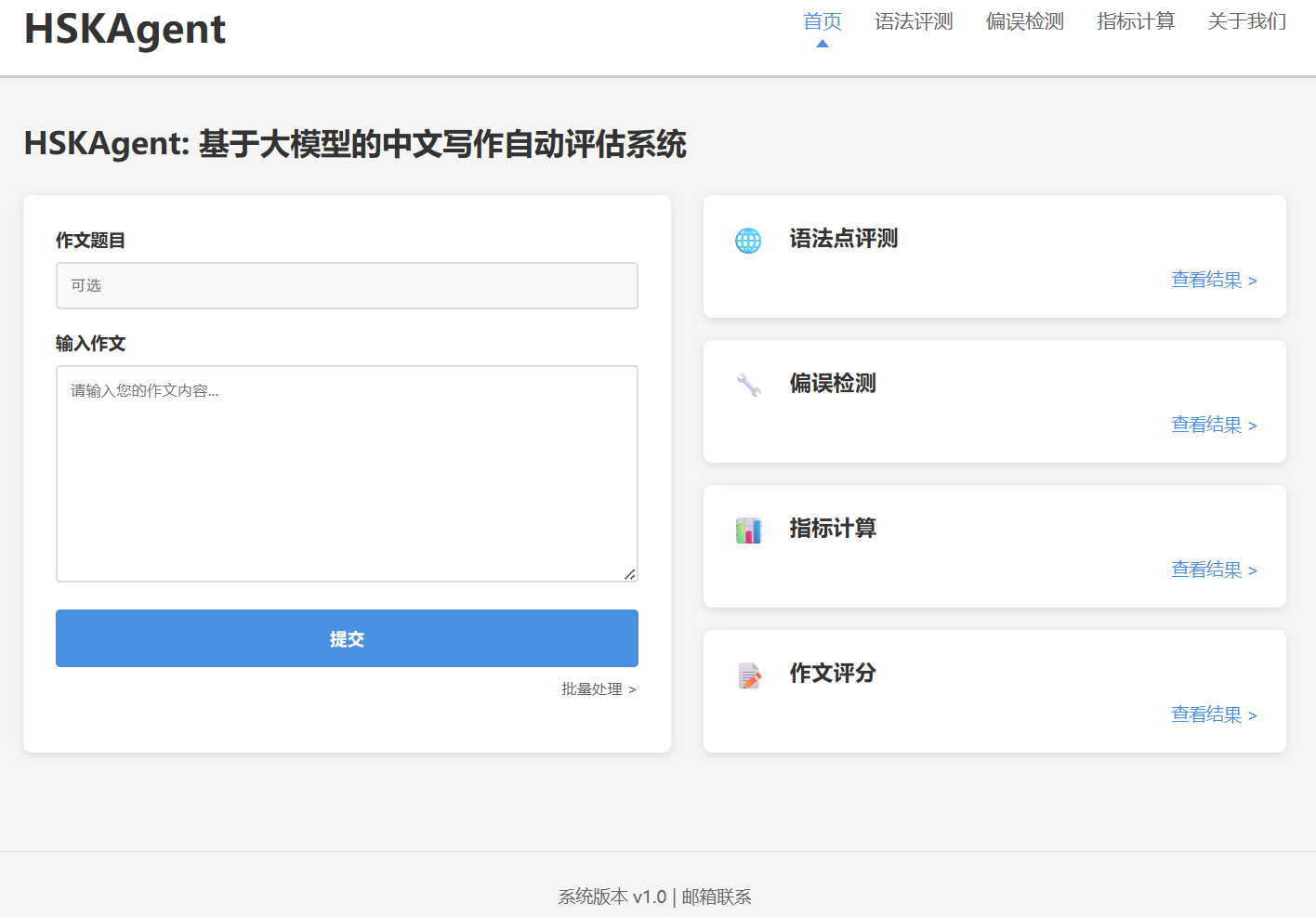}}
\caption{The homepage in the HSKAgent platform (Appendix G).}
\label{appendix_myfig2}
\end{figure}

\begin{figure}[H]
\centering
\fcolorbox{gray}{white}{\includegraphics[width=0.7\textwidth]{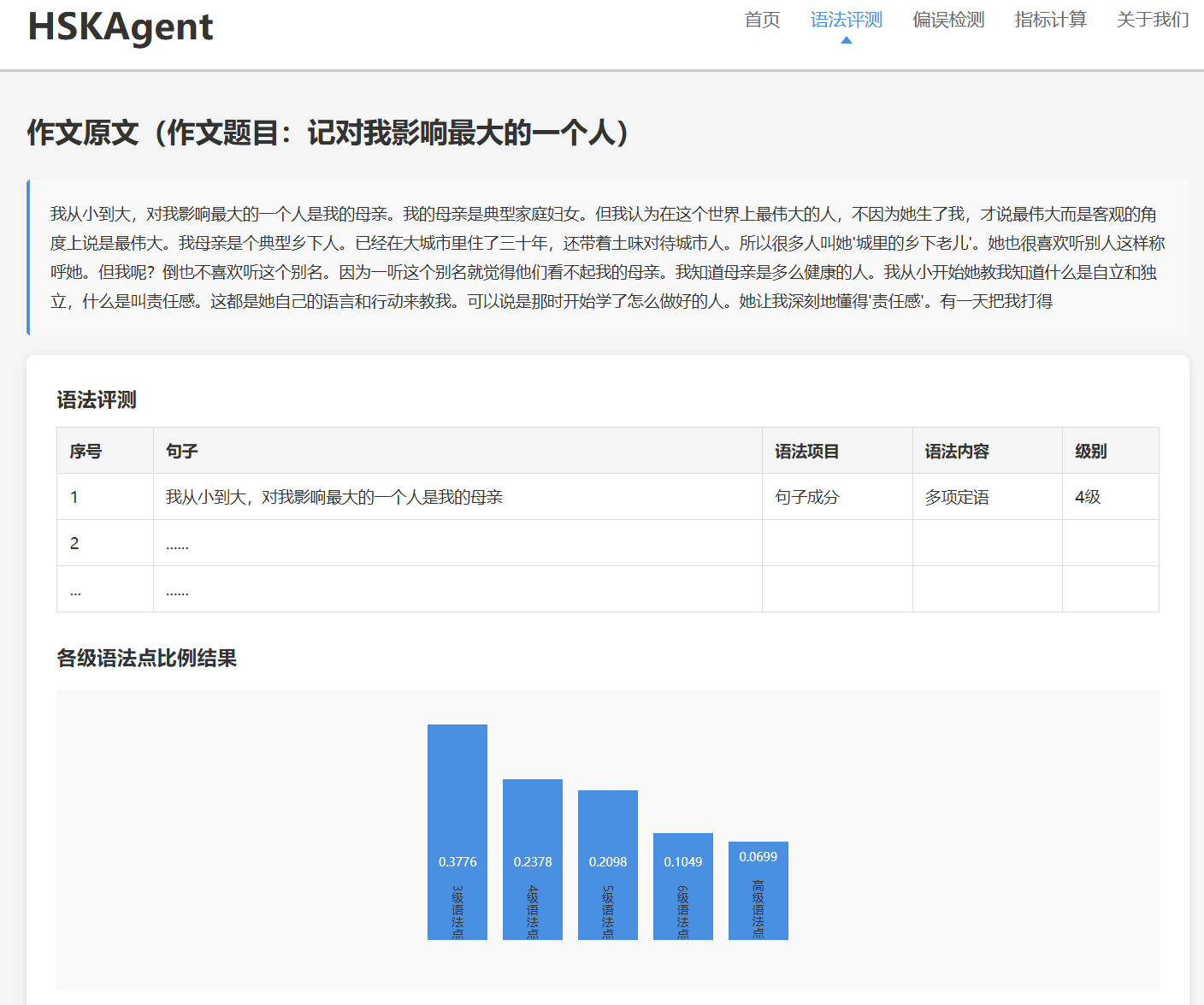}}
\caption{The function of the assessment of grammar items in the HSKAgent platform (Appendix G).}
\label{appendix_myfig3}
\end{figure}

\begin{figure}[H]
\centering
\fcolorbox{gray}{white}{\includegraphics[width=0.7\textwidth]{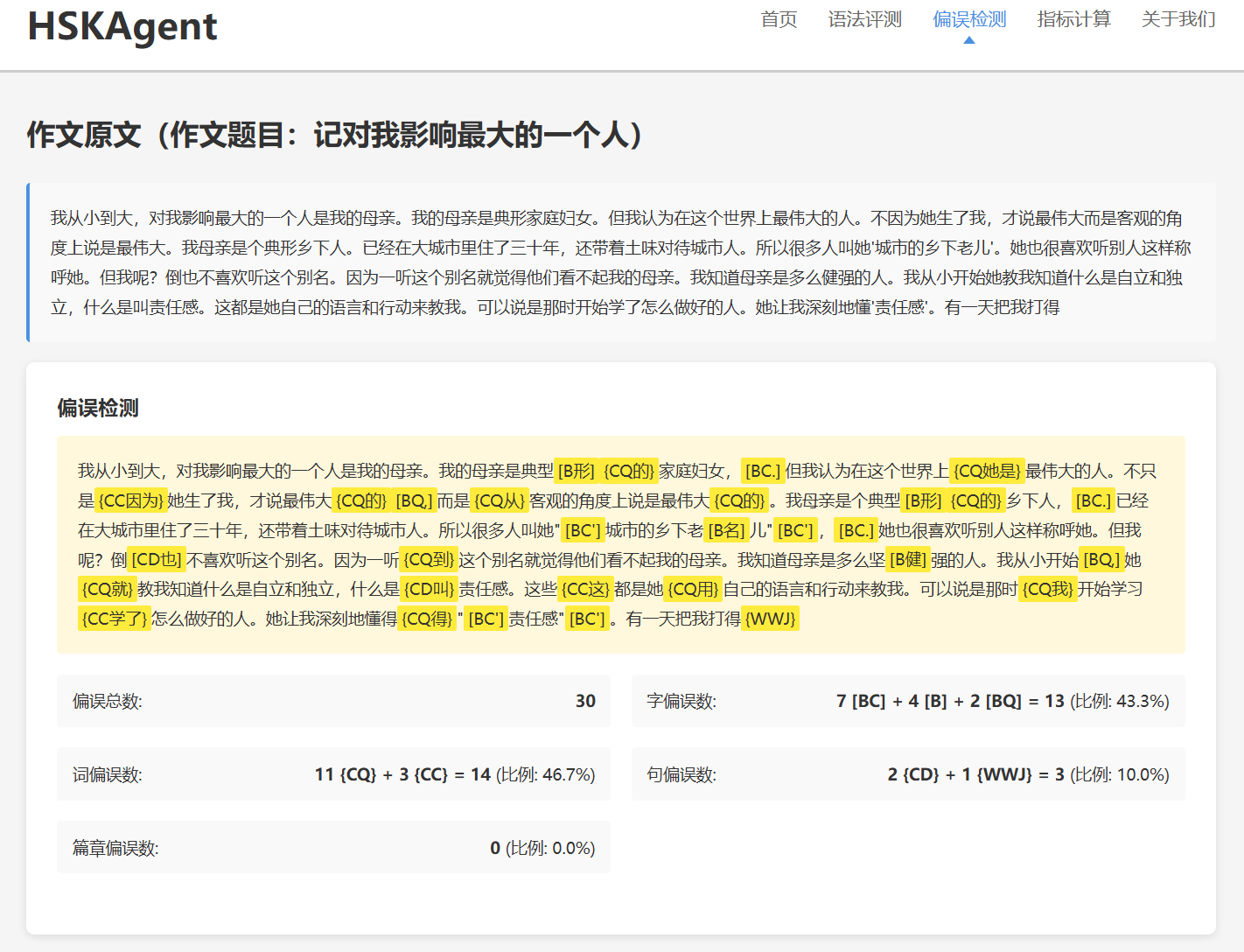}}
\caption{The function of error detection in the HSKAgent platform (Appendix G).}
\label{appendix_myfig4}
\end{figure}

\begin{figure}[H]
\centering
\fcolorbox{gray}{white}{\includegraphics[width=0.7\textwidth]{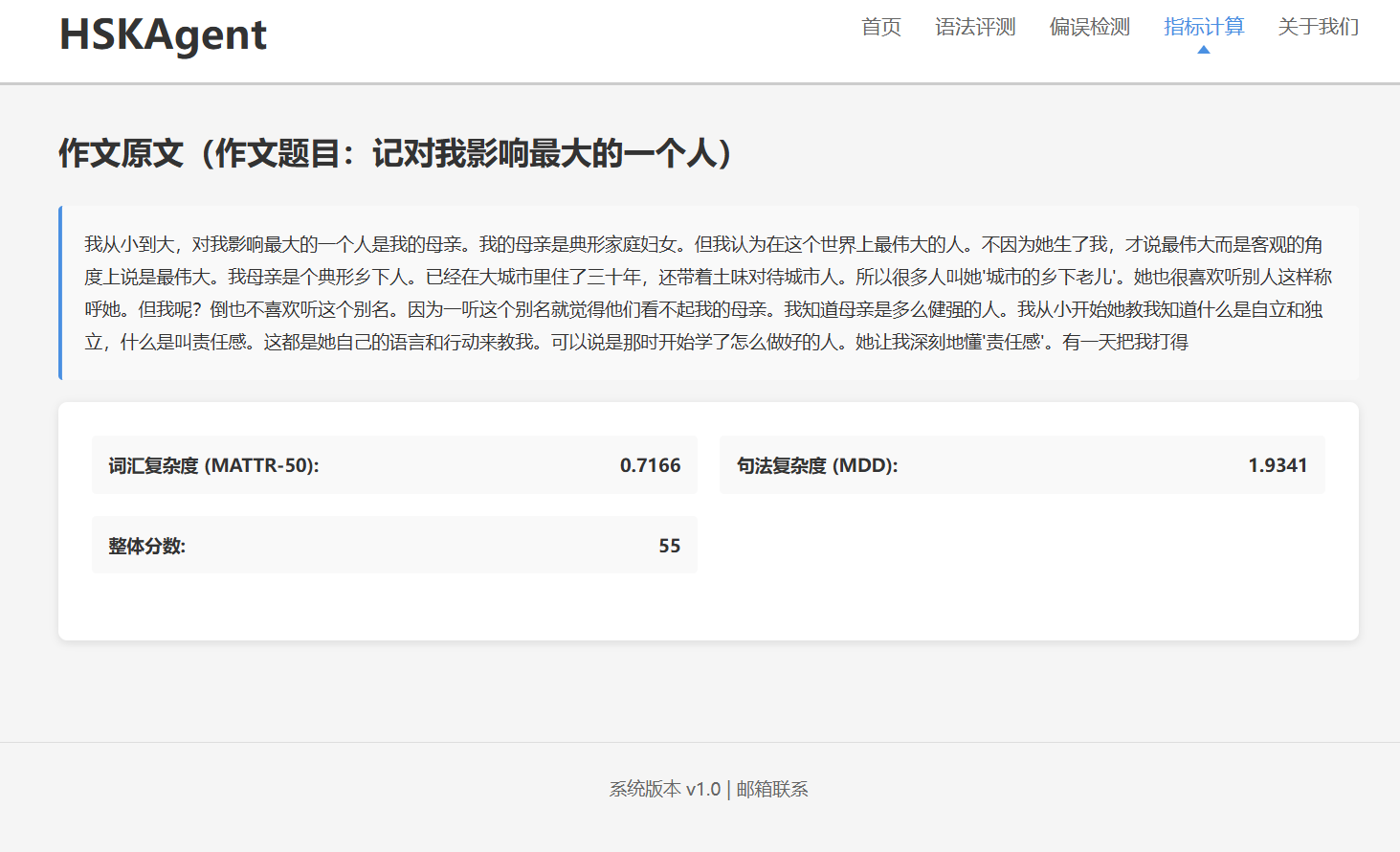}}
\caption{The functions of metric calculation and holistic scoring in the HSKAgent platform (Appendix G).}
\label{appendix_myfig5}
\end{figure}

To use this HSKAgent efficiently to assess the Chinese SLA development in LLMs, HSKAgent is deployed on a platform. The functional diagram of the platform is shown in Figure \ref{appendix_myfig2} to \ref{appendix_myfig5}. In the near future, we plan to launch an online version of this platform, aiming to provide professional and accurate writing assessment services to a wide range of Chinese learners.

\begin{CJK}{UTF8}{gbsn}       

\begin{table*}[hb!]
\centering
\begin{tabular}{lc|lc}
    \toprule
    \multicolumn{4}{c}{\textbf{Textbook Checklist (HSK level 3 to 4)}} \\
    \midrule
    \multicolumn{2}{c|}{\textbf{HSK 3}} & \multicolumn{2}{c}{\textbf{HSK 4}} \\
    \makecell[c]{Titles} & Tokens &
    \makecell[c]{Titles} & Tokens \\
    \midrule
\makecell[l]{考试大纲 HSK3 \\ \textit{Exam Outline HSK3}} & 12798 & \makecell[l]{考试大纲 HSK4 \\ \textit{Exam Outline HSK3}} & 14017 \\
\makecell[l]{词汇 HSK3 \\ \textit{Vocabulary HSK3}} & 9931 & \makecell[l]{词汇 HSK4 \\ \textit{Vocabulary HSK4}} & 20844 \\
\makecell[l]{教师用书 HSK3 \\ \textit{HSK3 Teacher's Book}} & 109898 & \makecell[l]{教师用书HSK4上 \\ \textit{HSK4 Teacher's Book 01}} & 78372 \\
\makecell[l]{标准教程3 HSK \\ \textit{HSK Standard Course 3}} & 49825 & \makecell[l]{教师用书HSK4下 \\ \textit{HSK4 Teacher's Book 02}} & 89045 \\
\makecell[l]{中文4 \\ \textit{Chinese 4}} & 23372 & \makecell[l]{HSK标准教程4上 \\ \textit{HSK Standard Course 4 01}} & 49702 \\
\makecell[l]{中文5 \\ \textit{Chinese 5}} & 29701 & \makecell[l]{HSK标准教程4下 \\ \textit{HSK Standard Course 4 02}} & 60742 \\
\makecell[l]{中文6 \\ \textit{Chinese 6}} & 35980 & \makecell[l]{中文7 \\ \textit{Chinese 7}} & 34967 \\
\makecell[l]{初级综合-01 \\ \textit{Elementary Comprehensive-01}} & 81155 & \makecell[l]{中文8 \\ \textit{Chinese 8}} & 38897 \\
\makecell[l]{初级综合-02 \\ \textit{Elementary Comprehensive-02}} & 108997 & \makecell[l]{中级写作-01 \\ \textit{Intermediate Writing-01}} & 46656 \\
\makecell[l]{初级读写-01 \\ \textit{Elementary Reading and Writing-01}} & 39899 & \makecell[l]{中级写作-02 \\ \textit{Intermediate Writing-02}} & 49390 \\
\makecell[l]{初级读写-02 \\ \textit{Elementary Reading and Writing-02}} & 30140 & \makecell[l]{中级综合-01 \\ \textit{Intermediate Comprehensive-01}} & 90602 \\
\makecell[l]{博雅汉语01 \\ \textit{Boya Chinese 01}} & 83216 & \makecell[l]{中级综合-02 \\ \textit{Intermediate Comprehensive-02}} & 157269 \\
\makecell[l]{博雅汉语02 \\ \textit{Boya Chinese 02}} & 69563 & \makecell[l]{中级阅读-01 \\ \textit{Intermediate Reading-01}} & 82834 \\
\makecell[l]{新HSK词汇突破3级 \\ \textit{New HSK Vocabulary Breakthrough HSK3}} & 29645 & \makecell[l]{中级阅读-02 \\ \textit{Intermediate Reading-02}} & 83005 \\
\makecell[l]{语法点速记3-4级 \\ \textit{Grammar Item Shorthand Level 3-4}} & 180917 & \makecell[l]{阅读写作HSK4(刘云) \\ \textit{Reading and Writing HSK4 (Liuyun)}} & 61067 \\
~ & ~ & \makecell[l]{博雅汉语03 \\ \textit{Boya Chinese 03}} & 120001 \\
~ & ~ & \makecell[l]{博雅汉语04 \\ \textit{Boya Chinese 04}} & 47060 \\
~ & ~ & \makecell[l]{四级应试指南 \\ \textit{Level 4 Exam Guide}} & 140688 \\
~ & ~ & \makecell[l]{新HSK词汇突破4级 \\ \textit{New HSK Vocabulary Breakthrough HSK4}} & 27441 \\
~ & ~ & \makecell[l]{语法点速记3-4级 \\ \textit{Grammar Item Shorthand Level 3-4}} & 180917 \\
\midrule
Total & 895037 & Total & 1473516 \\

    \bottomrule
\end{tabular}
\caption{The checklist of textbooks with HSK level 3 to 4 (Appendix A).}
\label{appendix_table1}
\end{table*}

\begin{table*}[hb!]
\centering
\setlength{\tabcolsep}{4pt}
\begin{tabular}{lc|lc}
    \toprule
    \multicolumn{4}{c}{\textbf{Textbook Checklist (HSK level 5 to 6)}} \\
    \midrule
    \multicolumn{2}{c|}{\textbf{HSK 5}} & \multicolumn{2}{c}{\textbf{HSK 6}} \\
    \makecell[c]{Titles} & Tokens &
    \makecell[c]{Titles} & Tokens \\
    \midrule
\makecell[l]{考试大纲 HSK5 \\ \textit{Exam Outline HSK5}} & 21130 & \makecell[l]{21天征服6级阅读 \\ \textit{21Days Reading HSK6}} & 87960 \\
\makecell[l]{词汇 HSK5 \\ \textit{Vocabulary HSK5}} & 7973 & \makecell[l]{21天征服6级写作 \\ \textit{21Days Writing HSK6}} & 80642 \\
\makecell[l]{教师用书HSK5上 \\ \textit{HSK5 Teacher's Book 01}} & 93142 & \makecell[l]{考试大纲 HSK6 \\ \textit{Exam Outline HSK6 }} & 37425 \\
\makecell[l]{教师用书HSK5下 \\ \textit{HSK5 Teacher's Book 02}} & 91246 & \makecell[l]{词汇 HSK6 \\ \textit{Vocabulary HSK6}} & 54514 \\
\makecell[l]{HSK标准教程5上 \\ \textit{HSK Standard Course 5 01}} & 61539 & \makecell[l]{教师用书HSK6上 \\ \textit{HSK6 Teacher's Book 01}} & 175899 \\
\makecell[l]{HSK标准教程5下 \\ \textit{HSK Standard Course 5 02}} & 63188 & \makecell[l]{教师用书HSK6下 \\ \textit{HSK6 Teacher's Book 02}} & 190876 \\
\makecell[l]{HSK语法点速记速练(高级篇) \\ \textit{HSK Grammar Item Quick Practice (Advanced)}} & 156408 & \makecell[l]{HSK标准教程6上 \\ \textit{HSK Standard Course 6 01}} & 99274 \\
\makecell[l]{中文10 \\ \textit{Chinese 10}} & 51916 & \makecell[l]{HSK标准教程6下 \\ \textit{HSK Standard Course 6 02}} & 112459 \\
\makecell[l]{中文9 \\ \textit{Chinese 9}} & 47366 & \makecell[l]{中文11 \\ \textit{Chinese 11}} & 51470 \\
\makecell[l]{中级写作-01 \\ \textit{Intermediate Writing-01}} & 46656 & \makecell[l]{中文12 \\ \textit{Chinese 12}} & 59157 \\
\makecell[l]{中级写作-02 \\ \textit{Intermediate Writing-02}} & 49390 & \makecell[l]{六级应试指南 \\ \textit{Level 6 Exam Guide}} & 253116 \\
\makecell[l]{中级综合-01 \\ \textit{Intermediate Comprehensive-01}} & 90602 & \makecell[l]{写作 HSK6 (刘云) \\ \textit{Writing HSK6 (Liuyun)}} & 101735 \\
\makecell[l]{中级综合-02 \\ \textit{Intermediate Comprehensive-02}} & 157269 & \makecell[l]{阅读 HSK6 (刘云) \\ \textit{Reading HSK6 (Liuyun)}} & 191351 \\
\makecell[l]{中级阅读-01 \\ \textit{Intermediate Reading-01}} & 82834 & \makecell[l]{博雅汉语07 \\ \textit{Boya Chinese 07}} & 87443 \\
\makecell[l]{中级阅读-02 \\ \textit{Intermediate Reading-02}} & 83005 & \makecell[l]{博雅汉语08 \\ \textit{Boya Chinese 08}} & 164812 \\
\makecell[l]{五级应试指南 \\ \textit{Level 5 Exam Guide}} & 63365 & \makecell[l]{6级语法点(外研社) \\ \textit{Level 6 Grammar Items (Waiyanshe)}} & 172650 \\
\makecell[l]{写作 HSK5 (刘云) \\ \textit{Writing HSK5 (Liuyun)}} & 40196 & \makecell[l]{新HSK词汇突破6级 \\ \textit{New HSK Vocabulary Breakthrough Level 6}} & 134159 \\
\makecell[l]{阅读 HSK5 (刘云) \\ \textit{Reading HSK5 (Liuyun)}} & 147227 & \makecell[l]{高级写作-01 \\ \textit{Advanced Writing-01}} & 54769 \\
\makecell[l]{博雅汉语05 \\ \textit{Boya Chinese 05}} & 159425 & \makecell[l]{高级写作-02 \\ \textit{Advanced Writing-02}} & 61396 \\
\makecell[l]{博雅汉语06 \\ \textit{Boya Chinese 06}} & 139426 & \makecell[l]{高级综合-01 \\ \textit{Advanced Comprehensive-01}} & 130526 \\
\makecell[l]{新HSK词汇突破5级 \\ \textit{New HSK Vocabulary Breakthrough Level 5}} & 63875 & \makecell[l]{高级综合-02 \\ \textit{Advanced Comprehensive-02}} & 150498 \\
~ & ~ & \makecell[l]{高级阅读-01 \\ \textit{Advanced Reading-01}} & 106868 \\
~ & ~ & \makecell[l]{高级阅读-02 \\ \textit{Advanced Reading-02}} & 119622 \\
\midrule
Total & 1717178 & Total & 2678621 \\

    \bottomrule
\end{tabular}
\caption{The checklist of textbooks with HSK level 5 to 6 (Appendix A).}
\label{appendix_table2}
\end{table*}

\onecolumn
\begin{center}
\begin{longtable}{lcccc>{\raggedright\arraybackslash}p{6.5cm}} 
\toprule
\multicolumn{6}{c}{\textbf{Grammar Items (HSK level 3)}} \\
\midrule
\makecell[c]{\textbf{No.}} & \textbf{Level} & \textbf{Item} & \textbf{Type} & \textbf{Detail} & \makecell[c]{\textbf{Content}} \\
\midrule
\endfirsthead

\toprule
\multicolumn{6}{c}{\textbf{Grammar Items (HSK level 3)} \textit{(continued)}} \\
\midrule
\makecell[c]{\textbf{No.}} & \textbf{Level} & \textbf{Item} & \textbf{Type} & \textbf{Detail} & \makecell[c]{\textbf{Content}} \\
\midrule
\endhead

\midrule
\multicolumn{6}{r}{\textit{(continued on next page)}} \\
\midrule
\endfoot

\endlastfoot

133 & 三级 & 词类 & 动词 & 能愿动词 & 敢\\
134 & 三级 & 词类 & 动词 & 能愿动词 & 需要\\
135 & 三级 & 词类 & 动词 & 离合词（动宾式） & 帮忙、点头、放假、干杯、见面、结婚、看病、睡觉、洗澡、理发、说话\\
136 & 三级 & 词类 & 动词 & 离合词（动补式） & 打开、看见、离开、完成\\
137 & 三级 & 词类 & 代词 & 疑问代词 & 疑问代词的非疑问用法：（1）任指用法：疑问代词+都…… / 疑问代词……疑问代词……；\#（2）不定指用法\\
138 & 三级 & 词类 & 代词 & 指示代词 & 各、各种、各个、每、任何\\
139 & 三级 & 词类 & 量词 & 名量词 & 把、行、架、群、束、双、台、张、支、只、种\\
140 & 三级 & 词类 & 量词 & 动量词 & 顿、口、眼\\
141 & 三级 & 词类 & 量词 & 量词重叠 & 量词重叠：AA\\
142 & 三级 & 词类 & 副词 & 程度副词 & 比较、更、还、相当\\
143 & 三级 & 词类 & 副词 & 范围、协同副词 & 光、仅、仅仅、就、至少\\
144 & 三级 & 词类 & 副词 & 时间副词 & 本来、才、曾经、从来、赶紧、赶快、立刻、连忙、始终、已、早已\\
145 & 三级 & 词类 & 副词 & 频率、重复副词 & 通常、往往、总、总是\\
146 & 三级 & 词类 & 副词 & 关联副词 & 再\\
147 & 三级 & 词类 & 副词 & 方式副词 & 互相、尽量、亲自、相互\\
148 & 三级 & 词类 & 副词 & 情态副词 & 大概、恐怕\\
149 & 三级 & 词类 & 副词 & 语气副词 & 白、并、当然、到底、反正、根本、果然、简直、绝对、难道、其实、千万、确实、只好、终于\\
150 & 三级 & 词类 & 介词 & 引出时间、处所 & 由1\\
151 & 三级 & 词类 & 介词 & 引出时间、处所 & 自从\\
152 & 三级 & 词类 & 介词 & 引出方向、路径 & 朝\\
153 & 三级 & 词类 & 介词 & 引出对象 & 为\\
154 & 三级 & 词类 & 介词 & 引出对象 & 向\\
155 & 三级 & 词类 & 介词 & 引出目的、原因 & 由于、因为\\
156 & 三级 & 词类 & 介词 & 引出目的、原因 & 为了\\
157 & 三级 & 词类 & 介词 & 引出施事、受事 & 把、被、叫、让\\
158 & 三级 & 词类 & 介词 & 表示排除 & 除了\\
159 & 三级 & 词类 & 介词 & 引出凭借、依据 & 按、按照\\
160 & 三级 & 词类 & 连词 & 连接分句或句子 & 并且、不光、不仅、另外、要是、于是、因此、由于、只有\\
161 & 三级 & 词类 & 拟声词 & ~ & 哈哈\\
162 & 三级 & 短语 & 结构类型 & 其他结构类型 & 其他结构类型2：①介宾短语 \# ②方位短语\#  ③兼语短语\#  ④同位短语\\
163 & 三级 & 短语 & 结构类型 & ~ & 数量重叠：数词+量词+数词+量词\\
164 & 三级 & 短语 & 固定短语 & 四字格 & 不A不B\\
165 & 三级 & 短语 & 固定短语 & 其他 & 看起来\\
166 & 三级 & 短语 & 固定短语 & 其他 & 看上去\\
167 & 三级 & 短语 & 固定短语 & 其他 & 有的是\\
168 & 三级 & 固定格式 & ~ & ~ & 除了……（以外），……还/也/都……\\
169 & 三级 & 固定格式 & ~ & ~ & 从……起\\
170 & 三级 & 固定格式 & ~ & ~ & 对……来说\\
171 & 三级 & 固定格式 & ~ & ~ & 像……一样\\
172 & 三级 & 固定格式 & ~ & ~ & 越……越……\\
173 & 三级 & 句子成分 & 主语 & ~ & 动词或动词性短语作主语 \# 形容词或形容词性短语作主语\\
174 & 三级 & 句子成分 & 宾语 & ~ & 动词或动词性短语作宾语 \# 形容词或形容词性短语和主谓短语作宾语\\
175 & 三级 & 句子成分 & 定语 & ~ & 动词或动词性短语作定语 \# 主谓短语作定语\\
176 & 三级 & 句子成分 & 补语 & 结果补语 & 结果补语：动词+到/住/走\\
177 & 三级 & 句子成分 & 补语 & 趋向补语 & 复合趋向补语的趋向意义用法：动词+出来/出去/过去/过来/回来/进去/进来/起来/上来/上去/下来/下去\\
178 & 三级 & 句子成分 & 补语 & 可能补语 & 可能补语：动词+得/不+动词/形容词 \# 动词+得/不+了\\
179 & 三级 & 句子成分 & 补语 & 程度补语 & 程度补语：形容词/心理动词+得很/极了/死了\\
180 & 三级 & 句子成分 & 补语 & 数量补语 & 数量补语（动词+数量补语）：宾语和数量补语共现\\
181 & 三级 & 句子成分 & 补语 & 数量补语 & 数量补语（动词+时量补语）：表示动作持续的时间\\
182 & 三级 & 句子成分 & 补语 & 数量补语 & 数量补语（动词+时量补语）：表示动作结束后到某个时间点的间隔时间\\
183 & 三级 & 句子的类型 & 句型 & 单句 & 主谓句4：主谓谓语句\\
184 & 三级 & 句子的类型 & 特殊句型 & “把”字句 & “把”字句：表处置（1）主语+把+宾语+动词+在/到+处所 \# “把”字句：（2）主语+把+宾语1+动词（+给）+宾语2 \#“把”字句：（3）主语+把+宾语+动词+结果补语/趋向补语/状态补语\\
185 & 三级 & 句子的类型 & 特殊句型 & 被动句 & 被动句：主语+被/叫/让+宾语+动词+其他成分\\
186 & 三级 & 句子的类型 & 特殊句型 & 连动句 & 连动句：（1）前一动作是后一动作的方式 \#连动句：（2）后一动作是前一动作的目的\\
187 & 三级 & 句子的类型 & 特殊句型 & 兼语句 & 兼语句 表使令：主语+叫/派/请/让……+宾语1+动词+宾语2\\
188 & 三级 & 句子的类型 & 特殊句型 & 比较句 & 比较句：（1）A比B+动词+得+形容词 \#比较句：（2）A不比B+形容词\#比较句：（3）A+动词+得+比+B+形容词\#比较句：（4）A比B+多少/早/晚+动词+数量短语\\
189 & 三级 & 句子的类型 & 特殊句型 & 重动句 & 主语+动词+宾语+动词+补语\\
190 & 三级 & 句子的类型 & 复句 & 并列复句 & （也）……，也……\\
191 & 三级 & 句子的类型 & 复句 & 并列复句 & 一会儿……，一会儿……\\
192 & 三级 & 句子的类型 & 复句 & 并列复句 & 一方面……，另一方面……\\
193 & 三级 & 句子的类型 & 复句 & 并列复句 & 又……，又……\\
194 & 三级 & 句子的类型 & 复句 & 承接复句 & 首先……，然后……\\
195 & 三级 & 句子的类型 & 复句 & 递进复句 & ……，并且……\\
196 & 三级 & 句子的类型 & 复句 & 递进复句 & 不仅/不光……，还/而且……\\
197 & 三级 & 句子的类型 & 复句 & 选择复句 & 不是……，就是……\\
198 & 三级 & 句子的类型 & 复句 & 转折复句 & ……X是X，就是/不过……\\
199 & 三级 & 句子的类型 & 复句 & 假设复句 & 要是……，就……\\
200 & 三级 & 句子的类型 & 复句 & 条件复句 & 只有……，才……\\
201 & 三级 & 句子的类型 & 复句 & 因果复句 & （由于……，）所以/因此……\\
202 & 三级 & 句子的类型 & 复句 & 目的复句 & 为了……，……\\
203 & 三级 & 句子的类型 & 复句 & 紧缩复句 & ……了……（就）……\\
205 & 三级 & 强调的方法 & ~ & ~ & 用“一点儿也不……”表示强调\\
206 & 三级 & 强调的方法 & ~ & ~ & 用反问句表示强调 反问句1：不是……吗？/难道……吗？\\
207 & 三级 & 强调的方法 & ~ & ~ & 用“是”表示强调\\
\bottomrule
\caption*{Table 6. The checklist of grammar items with HSK level 3 (Appendix B).}
\label{appendix_mytable3}
\end{longtable}

\begin{longtable}{lcccc>{\raggedright\arraybackslash}p{6.5cm}} 
\toprule
\multicolumn{6}{c}{\textbf{Grammar Items (HSK level 4)}} \\
\midrule
\makecell[c]{\textbf{No.}} & \textbf{Level} & \textbf{Item} & \textbf{Type} & \textbf{Detail} & \makecell[c]{\textbf{Content}} \\
\midrule
\endfirsthead

\toprule
\multicolumn{6}{c}{\textbf{Grammar Items (HSK level 4)} \textit{(continued)}} \\
\midrule
\makecell[c]{\textbf{No.}} & \textbf{Level} & \textbf{Item} & \textbf{Type} & \textbf{Detail} & \makecell[c]{\textbf{Content}} \\
\midrule
\endhead

\midrule
\multicolumn{6}{r}{\textit{(continued on next page)}} \\
\midrule
\endfoot

\endlastfoot

212 & 四级 & 词类 & 动词 & 能愿动词 & 得\\
213 & 四级 & 词类 & 代词 & 人称代词 & 人家\\
214 & 四级 & 词类 & 量词 & 名量词 & 打、袋、根、卷、裸、批\\
215 & 四级 & 词类 & 量词 & 借用量词 & （1）名量词：碗、脸、手、屋子、桌子；\#（2）动量词：刀、针\\
216 & 四级 & 词类 & 副词 & 程度副词 & 格外、极、极其\\
217 & 四级 & 词类 & 副词 & 范围、协同副词 & 共\\
218 & 四级 & 词类 & 副词 & 时间副词 & 按时、即将、急忙、渐渐、尽快\\
219 & 四级 & 词类 & 副词 & 频率、重复副词 & 一再、再三\\
220 & 四级 & 词类 & 副词 & 关联副词 & 却\\
221 & 四级 & 词类 & 副词 & 否定副词 & 未必\\
222 & 四级 & 词类 & 副词 & 情态副词 & 几乎、似乎\\
223 & 四级 & 词类 & 副词 & 语气副词 & 的确、反而、还、竟然、究竟\\
224 & 四级 & 词类 & 介词 & 引出时间、处所 & 自\\
225 & 四级 & 词类 & 介词 & 引出对象 & 对于\\
226 & 四级 & 词类 & 介词 & 引出对象 & 关于\\
227 & 四级 & 词类 & 介词 & 引出对象 & 替\\
228 & 四级 & 词类 & 介词 & 引出凭借、依据 & 根据\\
229 & 四级 & 词类 & 介词 & 引出凭借、依据 & 作为\\
230 & 四级 & 词类 & 连词 & 连接词或词组 & 并、以及\\
231 & 四级 & 词类 & 连词 & 连接分句或句子 & 此外、而、而是、既然、可见、甚至、假如、总之\\
232 & 四级 & 词类 & 助词 & 其他助词 & 似的\\
233 & 四级 & 词类 & 叹词 & ~ & 啊\\
234 & 四级 & 短语 & 固定短语 & 四字格 & 大A大B\\
235 & 四级 & 短语 & 固定短语 & 四字格 & 一A一B\\
236 & 四级 & 短语 & 固定短语 & 其他 & 看来\\
237 & 四级 & 短语 & 固定短语 & 其他 & 来得及/来不及\\
238 & 四级 & 短语 & 固定短语 & 其他 & 说不定\\
239 & 四级 & 短语 & 固定短语 & 其他 & 一般来说\\
240 & 四级 & 固定格式 & ~ & ~ & 一+量词+比+一+量词\\
241 & 四级 & 固定格式 & ~ & ~ & （自）……以来\\
242 & 四级 & 固定格式 & ~ & ~ & 由……组成\\
243 & 四级 & 固定格式 & ~ & ~ & 在……方面\\
244 & 四级 & 固定格式 & ~ & ~ & 在……上/下/中\\
245 & 四级 & 句子成分 & 主语 & 主谓短语作主语 & 主谓短语作主语\\
246 & 四级 & 句子成分 & 主语 & 受事主语 & 受事主语\\
247 & 四级 & 句子成分 & 定语 & 多项定语 & 多项定语\\
248 & 四级 & 句子成分 & 补语 & 趋向补语 & 趋向补语3 表示结果意义（引申用法）：动词+上/出/起/下\\
249 & 四级 & 句子的类型 & 特殊句型 & “把”字句 & “把”字句：表处置（1）主语+把+宾语+动词（+一/了）+动词 \#“把”字句（2）主语+把+宾语（+给）+动词+了/着 \#“把”字句（3）主语+把+宾语+动词+动量补语/时量补语\\
250 & 四级 & 句子的类型 & 特殊句型 & 被动句 & 被动句2：主语+被+动词+其他成分\\
251 & 四级 & 句子的类型 & 特殊句型 & 存现句 & 存现句2：（1）表示出现：处所词+动词+趋向补语/结果补语+动态助词（了）+数量短语+人物 \#存现句2：（2）表示消失：处所词+动词+结果补语+动态助词（了）+数量短语+人物\\
252 & 四级 & 句子的类型 & 特殊句型 & 兼语句 & 兼语句2（1）表爱惜义：主语+表扬/批评+宾语1+动词+宾语2 \#兼语句2（2）表称谓或认定义：主语+叫/称（呼）/说/认/选+宾语1+做/为/当/是+宾语2\\
253 & 四级 & 句子的类型 & 特殊句型 & “是……的”句 & “是……的”句2：强调说话人的看法或态度\\
254 & 四级 & 句子的类型 & 复句 & 并列复句 & 不是……，而是……\\
255 & 四级 & 句子的类型 & 复句 & 并列复句 & 既……，又/也……\\
256 & 四级 & 句子的类型 & 复句 & 承接复句 & 首先……，其次……\\
257 & 四级 & 句子的类型 & 复句 & 承接复句 & ……，于是……\\
258 & 四级 & 句子的类型 & 复句 & 递进复句 & ……，甚至……\\
259 & 四级 & 句子的类型 & 复句 & 选择复句 & 或者……，或者……\\
260 & 四级 & 句子的类型 & 复句 & 转折复句 & ……，然而……\\
261 & 四级 & 句子的类型 & 复句 & 假设复句 & ……，否则……\\
262 & 四级 & 句子的类型 & 复句 & 假设复句 & 假如……，(就)……\\
263 & 四级 & 句子的类型 & 复句 & 假设复句 & 万一……，(就)……\\
264 & 四级 & 句子的类型 & 复句 & 条件复句 & 不管……，都/也……\\
265 & 四级 & 句子的类型 & 复句 & 条件复句 & 无论……，都/也……\\
266 & 四级 & 句子的类型 & 复句 & 因果复句 & 既然……，就……\\
267 & 四级 & 句子的类型 & 复句 & 因果复句 & ……，可见……\\
268 & 四级 & 句子的类型 & 复句 & 让步复句 & 哪怕……，也/还……\\
269 & 四级 & 句子的类型 & 复句 & 目的复句 & ……，好……\\
270 & 四级 & 句子的类型 & 复句 & 紧缩复句 & 无标记\\
271 & 四级 & 句子的类型 & 复句 & 紧缩复句 & 不……也……\\
274 & 四级 & 强调的方法 & ~ & ~ & 用反问句表示强调 反问句2：由疑问代词构成的反问句\\
275 & 四级 & 强调的方法 & ~ & ~ & 用双重否定表示强调\\
276 & 四级 & 强调的方法 & ~ & ~ & 用“一+量词（+名词）+也（都）/也没（不）……”表示强调\\
277 & 四级 & 强调的方法 & ~ & ~ & 用“连……也/都……”表示强调\\
\bottomrule
\caption*{Table 7. The checklist of grammar items with HSK level 4 (Appendix B).}
\label{appendix_mytable4}
\end{longtable}

\newpage
\begin{longtable}{lcccc>{\raggedright\arraybackslash}p{6.5cm}} 
\toprule
\multicolumn{6}{c}{\textbf{Grammar Items (HSK level 5)}} \\
\midrule
\makecell[c]{\textbf{No.}} & \textbf{Level} & \textbf{Item} & \textbf{Type} & \textbf{Detail} & \makecell[c]{\textbf{Content}} \\
\midrule
\endfirsthead

\toprule
\multicolumn{6}{c}{\textbf{Grammar Items (HSK level 5)} \textit{(continued)}} \\
\midrule
\makecell[c]{\textbf{No.}} & \textbf{Level} & \textbf{Item} & \textbf{Type} & \textbf{Detail} & \makecell[c]{\textbf{Content}} \\
\midrule
\endhead

\midrule
\multicolumn{6}{r}{\textit{(continued on next page)}} \\
\midrule
\endfoot

\endlastfoot
288 & 五级 & 词类 & 代词 & 指示代词 & 彼此、如此\\
289 & 五级 & 词类 & 量词 & 名量词 & 册、朵、幅、届、颗、匹、扇\\
290 & 五级 & 词类 & 副词 & 程度副词 & 过于、可、稍、稍微、尤其\\
291 & 五级 & 词类 & 副词 & 范围、协同副词 & 大都\\
292 & 五级 & 词类 & 副词 & 时间副词 & 不时、将、将要、仍旧、时常、时刻、依旧、一向\\
293 & 五级 & 词类 & 副词 & 频率、重复副词 & 偶尔、再次\\
294 & 五级 & 词类 & 副词 & 方式副词 & 偷偷\\
295 & 五级 & 词类 & 副词 & 语气副词 & 毕竟、不免、差（一）点儿、倒是、干脆、就、居然、可、明明、总算\\
296 & 五级 & 词类 & 介词 & 引出时间、处所 & 随着\\
297 & 五级 & 词类 & 介词 & 引出目的、原因 & 将\\
298 & 五级 & 词类 & 介词 & 引出施事、受事 & 由\\
299 & 五级 & 词类 & 介词 & 引出凭借、依据 & 凭\\
300 & 五级 & 词类 & 介词 & 引出凭借、依据 & 依据\\
301 & 五级 & 词类 & 介词 & 引出凭借、依据 & 依照\\
302 & 五级 & 词类 & 介词 & 引出凭借、依据 & 依照\\
303 & 五级 & 词类 & 连词 & 连接分句或句子 & 从而、加上、完了、一旦\\
304 & 五级 & 词类 & 助词 & 其他助词 & 也好\\
305 & 五级 & 短语 & 固定短语 & 四字格 & A来A去\\
306 & 五级 & 短语 & 固定短语 & 四字格 & A着A着\\
307 & 五级 & 短语 & 固定短语 & 四字格 & 没A没B\\
308 & 五级 & 短语 & 固定短语 & 四字格 & 有A有B\\
309 & 五级 & 短语 & 固定短语 & 其他 & 不得了\\
310 & 五级 & 短语 & 固定短语 & 其他 & 不敢当\\
311 & 五级 & 短语 & 固定短语 & 其他 & 得了\\
312 & 五级 & 短语 & 固定短语 & 其他 & 用不着\\
313 & 五级 & 固定格式 & ~ & ~ & 从……来看\\
314 & 五级 & 固定格式 & ~ & ~ & 到……为止\\
315 & 五级 & 固定格式 & ~ & ~ & 够……的\\
316 & 五级 & 固定格式 & ~ & ~ & 拿……来说\\
317 & 五级 & 固定格式 & ~ & ~ & A的A，B的B\\
318 & 五级 & 固定格式 & ~ & ~ & 在……看来\\
319 & 五级 & 句子成分 & 宾语 & 宾语 & 宾语的语义类型1：（1）施事宾语；\# 宾语的语义类型1：（2）受事宾语\\
320 & 五级 & 句子成分 & 状语 & 多项状语 & 多项状语\\
321 & 五级 & 句子成分 & 补语 & 趋向补语 & 趋向补语4 表示时间意义（引申用法）（1）表示动作行为的开始：动词+上/起来 \#趋向补语（2）表示动作行为的持续：动词+下去/下来\\
322 & 五级 & 句子成分 & 补语 & 可能补语 & 可能补语2：动词+得/不得\\
323 & 五级 & 句子成分 & 补语 & 程度补语 & 程度补语2：（1）形容词/心理动词+得+不得了/慌/厉害； \#程度补语2：（2）动词/形容词+坏/透+了\\
324 & 五级 & 句子成分 & 补语 & 状态补语 & 状态补语2：动词/形容词+得+短语（1）动词/形容词+得+动词短语 \#状态补语2：（2）动词/形容词+主谓短语 \#状态补语2：（3）动词/形容词+得+固定短语\\
325 & 五级 & 句子的类型 & 特殊句型 & “有”字句 & “有”字句3：（1）表示存在、具有：主语+有+着+宾语； \#“有”字句3：（2）表示附着：主语+动词+有+宾语\\
326 & 五级 & 句子的类型 & 特殊句型 & “把”字句 & “把”字句3：表处置（1）主语+把+宾语+状语+动词 \#“把”字句3：（2）主语+把+宾语+一+动词 \#“把”字句3：（3）主语+把+宾语+动词+了 \#“把”字句3：（4）主语+把+宾语1+动词+宾语2\\
327 & 五级 & 句子的类型 & 特殊句型 & 被动句 & 被动句3：意念被动句\\
328 & 五级 & 句子的类型 & 特殊句型 & 连动句 & 连动句3：前后两个动词性词语具有因果、转折、条件关系\\
329 & 五级 & 句子的类型 & 特殊句型 & 兼语句 & 兼语句3 表致使：主语+叫/令/使/让+人称代词+动词短语\\
330 & 五级 & 句子的类型 & 特殊句型 & 比较句 & 比较句5：（1）跟……相比 \#比较句5： （2）A+形容词+B+数量补语\\
331 & 五级 & 句子的类型 & 复句 & 选择复句 & 或是……，或是……\\
332 & 五级 & 句子的类型 & 复句 & 转折复句 & 尽管……，但是/可是……\\
333 & 五级 & 句子的类型 & 复句 & 假设复句 & 一旦……，就……\\
334 & 五级 & 句子的类型 & 复句 & 假设复句 & 要是……（就）……，否则……\\
335 & 五级 & 句子的类型 & 复句 & 条件复句 & 除非……，才……\\
336 & 五级 & 句子的类型 & 复句 & 条件复句 & 除非……，否则/不然……\\
337 & 五级 & 句子的类型 & 复句 & 因果复句 & ……，因而……\\
338 & 五级 & 句子的类型 & 复句 & 让步复句 & 即使……，也……\\
339 & 五级 & 句子的类型 & 复句 & 目的复句 & ……，为的是……\\
340 & 五级 & 句子的类型 & 复句 & 目的复句 & ……，以便……\\
341 & 五级 & 句子的类型 & 复句 & 紧缩复句 & 没有……就没有……\\
342 & 五级 & 句子的类型 & 复句 & 紧缩复句 & 再……也……\\
343 & 五级 & 句子的类型 & 复句 & 多重复句 & 二重复句1：单句+复句；复句+单句\\
344 & 五级 & 强调的方法 & ~ & ~ & 用“再也不/没”表示强调\\
345 & 五级 & 强调的方法 & ~ & ~ & 用副词“可”表示强调\\
346 & 五级 & 强调的方法 & ~ & ~ & 用“怎么都/也+不/没”表示强调\\
\bottomrule
\caption*{Table 8. The checklist of grammar items with HSK level 5 (Appendix B).}
\label{appendix_mytable5}
\end{longtable}

\begin{longtable}{lcccc>{\raggedright\arraybackslash}p{6.5cm}} 
\toprule
\multicolumn{6}{c}{\textbf{Grammar Items (HSK level 6)}} \\
\midrule
\makecell[c]{\textbf{No.}} & \textbf{Level} & \textbf{Item} & \textbf{Type} & \textbf{Detail} & \makecell[c]{\textbf{Content}} \\
\midrule
\endfirsthead

\toprule
\multicolumn{6}{c}{\textbf{Grammar Items (HSK level 6)} \textit{(continued)}} \\
\midrule
\makecell[c]{\textbf{No.}} & \textbf{Level} & \textbf{Item} & \textbf{Type} & \textbf{Detail} & \makecell[c]{\textbf{Content}} \\
\midrule
\endhead

\midrule
\multicolumn{6}{r}{\textit{(continued on next page)}} \\
\midrule
\endfoot

\endlastfoot
361 & 六级 & 词类 & 代词 & 指示代词 & 本、此\\
362 & 六级 & 词类 & 量词 & 名量词 & 餐、串、滴、副、股、集、枝\\
363 & 六级 & 词类 & 量词 & 动量词 & 番、声、趟\\
364 & 六级 & 词类 & 副词 & 程度副词 & 特、异常\\
365 & 六级 & 词类 & 副词 & 范围、协同副词 & 尽、净、一齐、 一同\\
366 & 六级 & 词类 & 副词 & 时间副词 & 时时、一时、早晚\\
367 & 六级 & 词类 & 副词 & 关联副词 & 便\\
368 & 六级 & 词类 & 副词 & 方式副词 & 不禁、赶忙、亲眼、特地、特意\\
369 & 六级 & 词类 & 副词 & 情态副词 & 仿佛\\
370 & 六级 & 词类 & 副词 & 语气副词 & 才3、刚好、偏、恰好\\
371 & 六级 & 词类 & 介词 & 引出时间、处所 & 于\\
372 & 六级 & 词类 & 介词 & 引出方向、路径 & 沿（着）\\
373 & 六级 & 词类 & 介词 & 引出对象 & 同1、与1\\
374 & 六级 & 词类 & 介词 & 引出对象 & 至于\\
375 & 六级 & 词类 & 介词 & 引出目的、原因 & 因\\
376 & 六级 & 词类 & 介词 & 表示排除 & 除\\
377 & 六级 & 词类 & 介词 & 引出凭借、依据 & 据\\
378 & 六级 & 词类 & 连词 & 连接词或词组 & 而2、同2、与2\\
379 & 六级 & 词类 & 连词 & 连接分句或句子 & 不料、可3、若\\
380 & 六级 & 词类 & 助词 & 结构助词 & 所\\
381 & 六级 & 词类 & 助词 & 语气助词 & 罢了、啦、嘛\\
382 & 六级 & 短语 & 结构分类型 & 基本结构类型 & 数词+形容词+量词\\
383 & 六级 & 短语 & 固定短语 & 四字格 & 或A或B\\
384 & 六级 & 短语 & 固定短语 & 四字格 & 无A无B\\
385 & 六级 & 短语 & 固定短语 & 四字格 & A这A那\\
386 & 六级 & 短语 & 固定短语 & 四字格 & 左A右B\\
387 & 六级 & 短语 & 固定短语 & 其他 & 不怎么\\
388 & 六级 & 短语 & 固定短语 & 其他 & 不怎么样\\
389 & 六级 & 短语 & 固定短语 & 其他 & 好（不）容易\\
390 & 六级 & 短语 & 固定短语 & 其他 & 那倒（也）是\\
391 & 六级 & 短语 & 固定短语 & 其他 & 就是说/这就是说\\
392 & 六级 & 短语 & 固定短语 & 其他 & 算了\\
393 & 六级 & 固定格式 & 固定格式 & ~ & A→+量词，B→+量词\\
394 & 六级 & 固定格式 & 固定格式 & ~ & 东一A，西一A\\
395 & 六级 & 固定格式 & 固定格式 & ~ & 为了……而……\\
396 & 六级 & 句子成分 & 宾语 & 宾语 & 宾语的语义类型2：（1）处所宾语 \#宾语的语义类型2：（2）结果宾语\\
397 & 六级 & 句子成分 & 补语 & 趋向补语 & 趋向补语5 表示状态意义（引申用法）：动词/形容词+下来了/下去/起来/过来\\
398 & 六级 & 句子的类型 & 特殊句型 & “把”字句 & “把”字句4：表致使（1）主语（非生物体）+把+宾语+动词+其他成分 \#“把”字句4：（2）主语+把+宾语（施事）+动词+其他成分\\
399 & 六级 & 句子的类型 & 特殊句型 & 被动句 & 被动句4：主语+被/叫/让+宾语+给+动词+其他成分\\
400 & 六级 & 句子的类型 & 复句 & 并列复句 & 时而……，时而……\\
401 & 六级 & 句子的类型 & 复句 & 并列复句 & 一时……一时……\\
402 & 六级 & 句子的类型 & 复句 & 承接复句 & ……便……\\
403 & 六级 & 句子的类型 & 复句 & 递进复句 & 不但不/不但没有……，反而……\\
404 & 六级 & 句子的类型 & 复句 & 递进复句 & 不是……，还/还是……\\
405 & 六级 & 句子的类型 & 复句 & 递进复句 & 连……都/也……，……更……\\
406 & 六级 & 句子的类型 & 复句 & 选择复句 & 要么……，要么……\\
407 & 六级 & 句子的类型 & 复句 & 转折复句 & 虽……，但/可/却/也……\\
408 & 六级 & 句子的类型 & 复句 & 假设复句 & ……，要不然/不然……\\
409 & 六级 & 句子的类型 & 复句 & 条件复句 & 凡是……，都……\\
410 & 六级 & 句子的类型 & 复句 & 让步复句 & 就算/就是……也……\\
411 & 六级 & 句子的类型 & 复句 & 紧缩复句 & 不……不……\\
412 & 六级 & 句子的类型 & 复句 & 多重复句 & 二重复句2：复句+复句\\
413 & 六级 & 强调的方法 & ~ & ~ & 用“非……不可”表示强调\\
\bottomrule
\caption*{Table 9. The checklist of grammar items with HSK level 6 (Appendix B).}
\label{appendix_mytable6}
\end{longtable}

\newpage
\begin{longtable}{lcccc>{\raggedright\arraybackslash}p{6.5cm}} 
\toprule
\multicolumn{6}{c}{\textbf{Grammar Items (HSK advanced level)}} \\
\midrule
\makecell[c]{\textbf{No.}} & \textbf{Level} & \textbf{Item} & \textbf{Type} & \textbf{Detail} & \makecell[c]{\textbf{Content}} \\
\midrule
\endfirsthead

\toprule
\multicolumn{6}{c}{\textbf{Grammar Items (HSK advanced level)} \textit{(continued)}} \\
\midrule
\makecell[c]{\textbf{No.}} & \textbf{Level} & \textbf{Item} & \textbf{Type} & \textbf{Detail} & \makecell[c]{\textbf{Content}} \\
\midrule
\endhead

\midrule
\multicolumn{6}{r}{\textit{(continued on next page)}} \\
\midrule
\endfoot

\endlastfoot
426 & 高等 & 词类 & 动词 & 能愿动词 & 需\\
427 & 高等 & 词类 & 代词 & 疑问代词 & 何\\
428 & 高等 & 词类 & 代词 & 指示代词 & 该、另、兹\\
429 & 高等 & 词类 & 量词 & 名量词 & (1）栋、粒、枚、则、盖 \# (2）复合量词：人次\\
430 & 高等 & 词类 & 副词 & 关联副词 & 亦\\
431 & 高等 & 词类 & 介词 & 引出方向、路径 & 顺着\\
432 & 高等 & 词类 & 连词 & 连接词或词组 & 及\\
433 & 高等 & 词类 & 连词 & 连接分句或句子 & 继而、要不是\\
434 & 高等 & 词类 & 助词 & 结构助词 & 之\\
519 & 高等 & 词类 & 副词 & 程度副词 & 极为\\
520 & 高等 & 词类 & 副词 & 程度副词 & 尽\\
521 & 高等 & 词类 & 副词 & 程度副词 & 蛮\\
522 & 高等 & 词类 & 副词 & 程度副词 & 颇\\
523 & 高等 & 词类 & 副词 & 程度副词 & 稍稍\\
524 & 高等 & 词类 & 副词 & 程度副词 & 尤为\\
525 & 高等 & 词类 & 副词 & 程度副词 & 越发\\
528 & 高等 & 词类 & 副词 & 范围、协同副词 & 凡\\
529 & 高等 & 词类 & 副词 & 范围、协同副词 & 皆\\
530 & 高等 & 词类 & 副词 & 范围、协同副词 & 统统\\
531 & 高等 & 词类 & 副词 & 范围、协同副词 & 唯独\\
532 & 高等 & 词类 & 副词 & 方式副词 & 不由得\\
533 & 高等 & 词类 & 副词 & 方式副词 & 一连\\
534 & 高等 & 词类 & 副词 & 方式副词 & 顺便\\
535 & 高等 & 词类 & 副词 & 否定副词 & 未\\
536 & 高等 & 词类 & 副词 & 否定副词 & 勿\\
537 & 高等 & 词类 & 副词 & 频率、重复副词 & 频频\\
538 & 高等 & 词类 & 副词 & 频率、重复副词 & 再度\\
543 & 高等 & 词类 & 副词 & 情态副词 & 必定\\
544 & 高等 & 词类 & 副词 & 情态副词 & 不妨\\
545 & 高等 & 词类 & 副词 & 情态副词 & 何必\\
546 & 高等 & 词类 & 副词 & 情态副词 & 莫非\\
547 & 高等 & 词类 & 副词 & 情态副词 & 按说\\
548 & 高等 & 词类 & 副词 & 时间副词 & 即\\
549 & 高等 & 词类 & 副词 & 时间副词 & 历来\\
550 & 高等 & 词类 & 副词 & 时间副词 & 尚\\
551 & 高等 & 词类 & 副词 & 时间副词 & 向来\\
552 & 高等 & 词类 & 介词 & 引出对象 & 当着\\
553 & 高等 & 词类 & 介词 & 引出对象 & 就5\\
554 & 高等 & 词类 & 介词 & 引出凭借、依据 & 趁\\
555 & 高等 & 词类 & 介词 & 引出凭借、依据 & 基于\\
556 & 高等 & 词类 & 介词 & 引出凭借、依据 & 依\\
557 & 高等 & 词类 & 副词 & 语气副词 & 白白\\
558 & 高等 & 词类 & 副词 & 语气副词 & 反倒\\
559 & 高等 & 词类 & 副词 & 语气副词 & 分明\\
560 & 高等 & 词类 & 副词 & 语气副词 & 怪不得\\
561 & 高等 & 词类 & 副词 & 语气副词 & 好在\\
562 & 高等 & 词类 & 副词 & 语气副词 & 乃\\
563 & 高等 & 词类 & 副词 & 语气副词 & 难怪\\
564 & 高等 & 词类 & 副词 & 语气副词 & 偏偏\\
565 & 高等 & 词类 & 副词 & 语气副词 & 索性\\
566 & 高等 & 词类 & 副词 & 语气副词 & 万万\\
567 & 高等 & 词类 & 副词 & 语气副词 & 未免\\
568 & 高等 & 词类 & 副词 & 语气副词 & 无非\\
569 & 高等 & 词类 & 副词 & 语气副词 & 幸好\\
570 & 高等 & 词类 & 副词 & 语气副词 & 幸亏\\
571 & 高等 & 词类 & 副词 & 语气副词 & 终究\\
572 & 高等 & 词类 & 助词 & 语气助词 & 而已\\
573 & 高等 & 词类 & 助词 & 语气助词 & 矣\\
435 & 高等 & 短语 & 结构类型 & 基本结构类型 & 数词+量词+抽象事物\\
436 & 高等 & 短语 & 固定短语 & 四字格 & 爱A不A\\
437 & 高等 & 短语 & 固定短语 & 四字格 & 半A半B\\
438 & 高等 & 短语 & 固定短语 & 四字格 & 东A西B\\
439 & 高等 & 短语 & 固定短语 & 四字格 & 非A非B\\
440 & 高等 & 短语 & 固定短语 & 四字格 & 忽A忽B\\
441 & 高等 & 短语 & 固定短语 & 四字格 & 连A带B\\
442 & 高等 & 短语 & 固定短语 & 四字格 & 时A时B\\
443 & 高等 & 短语 & 固定短语 & 四字格 & 自A自B\\
444 & 高等 & 短语 & 固定短语 & 其他 & 巴不得\\
445 & 高等 & 短语 & 固定短语 & 其他 & 别提了\\
446 & 高等 & 短语 & 固定短语 & 其他 & 除此之外\\
447 & 高等 & 短语 & 固定短语 & 其他 & 归根到底\\
448 & 高等 & 短语 & 固定短语 & 其他 & 可不是\\
449 & 高等 & 短语 & 固定短语 & 其他 & 没说的\\
450 & 高等 & 短语 & 固定短语 & 其他 & 无论如何\\
451 & 高等 & 短语 & 固定短语 & 其他 & 由此可见\\
539 & 高等 & 短语 & 固定短语 & 其他 & 与此同时\\
540 & 高等 & 短语 & 固定短语 & 其他 & 这样一来\\
541 & 高等 & 短语 & 固定短语 & 其他 & 综上所述\\
542 & 高等 & 短语 & 固定短语 & 其他 & 总的来说/总而言之\\
452 & 高等 & 固定格式 & ~ & ~ & 不知……好\\
453 & 高等 & 固定格式 & ~ & ~ & 所谓……就是……\\
454 & 高等 & 固定格式 & ~ & ~ & 无非/不过/只不过/只是……而已/罢了\\
455 & 高等 & 固定格式 & ~ & ~ & 以……为……\\
456 & 高等 & 固定格式 & ~ & ~ & 因……而……\\
457 & 高等 & 句子成分 & 宾语 & 宾语的语义类型3： & （1）方式宾语 \#（2）工具宾语 \#（3）材料宾语 \#（4）目的宾语\\
458 & 高等 & 句子成分 & 补语 & 程度补语3： & （1）形容词/动词+得+不行 \#（2）形容词/动词+得+要命/要死\\
459 & 高等 & 句子成分 & 补语 & 状态补语： & “个”引导的补语\\
460 & 高等 & 句子的类型 & 特殊句型 & “把”字句5： & 表致使（主语+）把+宾语（施事）+动词+了\\
461 & 高等 & 句子的类型 & 特殊句型 & 被动句 & （1）被……所…… \#（2）为……所……\\
462 & 高等 & 句子的类型 & 特殊句型 & 比较句6： & （1）比起……（来） \#（2）A+形容词+于+B \#（3）A+比+名词+还+名词\\
463 & 高等 & 句子的类型 & 复句 & 并列复句 & 一面……，一面……\\
464 & 高等 & 句子的类型 & 复句 & 承接复句 & ……，此后……\\
465 & 高等 & 句子的类型 & 复句 & 承接复句 & 起初……，……才……\\
466 & 高等 & 句子的类型 & 复句 & 递进复句 & 别说……，连……也/都……  \# 连……也/都……，别说…… \# 别说……，即使……也…… \# 即使……也……，别说……\\
467 & 高等 & 句子的类型 & 复句 & 递进复句 & ……，况且……\\
468 & 高等 & 句子的类型 & 复句 & 递进复句 & 连……，更不用说……\\
469 & 高等 & 句子的类型 & 复句 & 递进复句 & ……，乃至……\\
470 & 高等 & 句子的类型 & 复句 & 递进复句 & ……，且……\\
471 & 高等 & 句子的类型 & 复句 & 递进复句 & ……，甚至于……\\
472 & 高等 & 句子的类型 & 复句 & 选择复句 & 或……，或……\\
473 & 高等 & 句子的类型 & 复句 & 选择复句 & 宁可/宁愿……，也……\\
474 & 高等 & 句子的类型 & 复句 & 选择复句 & 与其……，不如……\\
475 & 高等 & 句子的类型 & 复句 & 选择复句 & 与其……，宁可/宁愿……\\
476 & 高等 & 句子的类型 & 复句 & 转折复句 & ……，而……（则）……\\
477 & 高等 & 句子的类型 & 复句 & 转折复句 & ……，……倒/反倒……\\
478 & 高等 & 句子的类型 & 复句 & 假设复句 & 倘若/若……，……\\
479 & 高等 & 句子的类型 & 复句 & 假设复句 & 倘若/假设/假使/若……，就/那么……\\
480 & 高等 & 句子的类型 & 复句 & 假设复句 & 幸亏……，要不然/不然/要不/否则……\\
481 & 高等 & 句子的类型 & 复句 & 条件复句 & 别管……，都……\\
482 & 高等 & 句子的类型 & 复句 & 条件复句 & 任……，也……\\
483 & 高等 & 句子的类型 & 复句 & 因果复句 & （因）……，故……\\
484 & 高等 & 句子的类型 & 复句 & 因果复句 & 鉴于……，……\\
485 & 高等 & 句子的类型 & 复句 & 因果复句 & （由于）……，以致……\\
486 & 高等 & 句子的类型 & 复句 & 因果复句 & ……，以至于……\\
487 & 高等 & 句子的类型 & 复句 & 因果复句 & 之所以……，是因为/是由于……\\
488 & 高等 & 句子的类型 & 复句 & 让步复句 & 固然……，也……\\
489 & 高等 & 句子的类型 & 复句 & 让步复句 & ……固然……，但是/可是/不过……\\
490 & 高等 & 句子的类型 & 复句 & 让步复句 & 即便……，也……\\
491 & 高等 & 句子的类型 & 复句 & 让步复句 & 虽说……，但是/可是/不过……\\
492 & 高等 & 句子的类型 & 复句 & 让步复句 & 纵然……，也……\\
493 & 高等 & 句子的类型 & 复句 & 目的复句 & ……，以……\\
494 & 高等 & 句子的类型 & 复句 & 目的复句 & ……，以免/免得……\\
495 & 高等 & 句子的类型 & 复句 & 解说复句 & ……，也就是说……\\
496 & 高等 & 句子的类型 & 复句 & 紧缩复句 & (要+) 动词+就+动词+个+补语\\
497 & 高等 & 句子的类型 & 复句 & 紧缩复句 & 动词（+宾语1）+就+动词（+宾语1）, 别……\\
498 & 高等 & 句子的类型 & 复句 & 多重复句 & 三重或三重以上的复句\\
526 & 高等 & 句子的类型 & 复句 & 递进复句 & ……，何况……\\
527 & 高等 & 句子的类型 & 复句 & 递进复句 & ……，进而……\\
499 & 高等 & 强调的方法 & ~ & ~ & 用反问句表示强调 反问句3：何必/何苦……呢？\\

\bottomrule
\caption*{Table 10. The checklist of grammar items with HSK advanced level (Appendix B).}
\label{appendix_mytable7}
\end{longtable}
\end{center}

\newpage

\begin{tcolorbox}[colback=gray!5!white,colframe=gray!75!black,breakable,enhanced,title=The prompt for generating instruction data based on grammar items.]
\textcolor{red}{\underline{\#\# [ROLE]}}
\vspace{\baselineskip} 

你是一位经验丰富的中文写作教学专家，擅长设计写作型语法教学任务。
\vspace{\baselineskip} 

\textcolor{red}{\underline{\#\# [Requirements]}}
\vspace{\baselineskip} 

我将会提供给你一个语法项目的结构化描述，请你根据这个语法项目，完成以下任务：
\vspace{\baselineskip} 

①请理解"语法项目"、"类别"、"细目"的可能的教学意图，将语法内容作为一个单独的练习对象，设计成一个用于学生写作练习的教学指令数据。  
\vspace{\baselineskip} 

②请你根据不同的HSK等级，设计不同难度的指令任务，旨在指引学生写一篇“XXX”，也可以是写一句/一段“XXX”。
\vspace{\baselineskip} 

③每个语法内容都可能包含不同的用法，对于此情况你可以用不同的用法来构造多样的教学指令任务。
\vspace{\baselineskip} 

④由于每个语法项目都需要生成若干条不同的教学指令数据，请你每一次输出新的教学指令数据时，不同与指令池里已有的教学指令重复。指令池已有内容： [Instruction Pool]
\vspace{\baselineskip} 

⑤你生成的指令数据必须以json格式输出。
\vspace{\baselineskip} 

\textcolor{red}{\underline{\#\# [Examples]}}
\vspace{\baselineskip} 

例子1：
\vspace{\baselineskip} 

下面是一个语法项目的结构化描述：
\vspace{\baselineskip} 

\{
\vspace{\baselineskip} 

\hspace*{2em}"HSK等级": "3级",
  \vspace{\baselineskip} 

\hspace*{2em}"语法项目": "词类",
\vspace{\baselineskip} 

\hspace*{2em}"类别": "动词",
\vspace{\baselineskip} 

\hspace*{2em}"细目": "离合词（动宾式）",
\vspace{\baselineskip} 

\hspace*{2em}"语法内容": "见面"
\vspace{\baselineskip} 

\}
\vspace{\baselineskip} 

你的输出 (json)：
\vspace{\baselineskip} 

\{
\vspace{\baselineskip} 

\hspace*{2em}"instruction": "请描述你上一次和朋友见面的经历，要求使用‘见面’这个词。",
\vspace{\baselineskip} 

\hspace*{2em}"input": "语法点：离合词（动宾式）；类别：动词；词条：见面",
\vspace{\baselineskip} 

\hspace*{2em}"output": "上个周末我和老同学在市中心见面，我们聊了很多高中时的趣事，还一起吃了火锅。"
\vspace{\baselineskip} 

\}
\vspace{\baselineskip} 

例子2：
\vspace{\baselineskip} 

下面是一个语法项目的结构化描述：
\vspace{\baselineskip} 

\{\vspace{\baselineskip} 

\hspace*{2em}"HSK等级": "3级",
\vspace{\baselineskip} 

\hspace*{2em}"语法项目": "句子成分", 
\vspace{\baselineskip} 

  \hspace*{2em}"类别": "补语", 
\vspace{\baselineskip} 

  \hspace*{2em}"细目": "可能补语", 
\vspace{\baselineskip} 

  \hspace*{2em}"语法内容": "可能补语：动词+得/不+动词/形容词" 
\vspace{\baselineskip} 

\} 
\vspace{\baselineskip} 

你的输出 (json)： 
\vspace{\baselineskip} 

\{ 
\vspace{\baselineskip} 

  \hspace*{2em}"instruction": "请写一段关于你尝试学习一项新技能的经历，并在描述中使用“动词+得/不+形容词”这一可能补语结构，表达你当时是否学得会、做得好。", 
\vspace{\baselineskip} 

  \hspace*{2em}"input": "语法点：可能补语；类别：补语；结构：动词+得/不+动词/形容词", 
\vspace{\baselineskip} 

  \hspace*{2em}"output": "去年我开始学习打羽毛球。一开始我动作做得不标准，总是接不到球。但我每天坚持练习，后来发球发得越来越好，跑动也变得很灵活。虽然比赛还赢不了，但教练说我学得挺快。" 
\vspace{\baselineskip} 

\} 
\vspace{\baselineskip} 

\textcolor{red}{\underline{\#\# [Task Descriptions]}}
\vspace{\baselineskip} 

现在，我将会提供给你一个新的语法项目，请你遵循json格式输出1条指令数据，最后的keys只包括 "instruction"、"input"、"output"。其中， 指令任务可以指导学生写“一篇/一段/一句”某个主题的内容，但是要确保"output"内容的词数要符合这个指令的要求。请注意：你生成新的教学指令数据时，不能与指令池里的已有指令重复，尽可能变化多样（不同场景，不同指令风格，不同要求）去设计这样的教学指令，所有内容都必须是中文。
\vspace{\baselineskip} 

新的语法项目：
\vspace{\baselineskip} 

[New Grammar Point]
\vspace{\baselineskip} 

你的输出 (json)：

\end{tcolorbox}
\begin{figure}[ht]
\centering
\caption{The prompt of generating instruction data based on the level-based grammar items (Appendix C).}
\label{appendix_myfig1}
\end{figure}

\end{CJK}

\newpage

\end{document}